\documentclass[lettersize,journal]{IEEEtran}
\usepackage{amsmath,amsfonts}
\usepackage{algorithmic}
\usepackage{algorithm}
\usepackage{array}
\usepackage[caption=false,font=normalsize,labelfont=sf,textfont=sf]{subfig}
\usepackage{textcomp}
\usepackage{stfloats}
\usepackage{url}
\usepackage{verbatim}
\usepackage{graphicx}
\usepackage{cite}
\usepackage[table]{xcolor} 
\usepackage{booktabs} 

\usepackage[top=0.55in, bottom=0.6in, left=0.5in, right=0.5in]{geometry}

\makeatletter
\g@addto@macro\normalsize{%
  \setlength\abovedisplayskip{3pt plus 2pt minus 2pt}
  \setlength\belowdisplayskip{3pt plus 2pt minus 2pt}
  \setlength\abovedisplayshortskip{1pt plus 1pt}
  \setlength\belowdisplayshortskip{2pt plus 2pt minus 1pt}
}
\makeatother

\definecolor{firstcolor}{HTML}{FDAE6B}
\definecolor{secondcolor}{HTML}{FDD0A2}
\definecolor{thirdcolor}{HTML}{FEE6CE}
\definecolor{restcolor}{HTML}{FFF5EB}

\begin{document}

% --- 论文标题 ---
\title{Continuous Splatting meets Retinex: Continuous Gaussian Splatting and Implicit Reflectance Modeling for Low-Light Image Enhancement}

% --- 作者与单位信息 ---
\author{Yuhan~Chen, 
        Yicui~Shi, 
        Guofa~Li*, 
        Wenxuan~Yu, 
        Ying~Fang, 
        Guangrui~Bai, 
        Wenbo~Chu, 
        and~Keqiang~Li%
\thanks{This work was supported by the National Natural Science Foundation of China under Grant No. 52272421. (Corresponding author: Guofa Li).}%
\thanks{Yuhan~Chen, Yicui~Shi, Guofa~Li, Wenxuan~Yu, and Ying~Fang are with the College of Mechanical and Vehicle Engineering, Chongqing University, Chongqing 400044, China(e-mail: 20240701028@stu.cqu.edu.cn; yicuishi@cqu.edu.cn; liguofa@cqu.edu.cn; wenxuanyu@cqu.edu.cn; yingfang@stu.cqu.edu.cn).}%
\thanks{Guangrui~Bai is with the School of Engineering Science, University of Science and Technology of China, Hefei 230026, China(e-mail: mubai@mail.ustc.edu).}%
\thanks{Wenbo~Chu is with the National Innovation Center of Intelligent and Connected Vehicles, Beijing 100089, China(e-mail: chuwenbo@wicv.cn).}%
\thanks{Keqiang~Li is with the School of Vehicle and Mobility, Tsinghua University, Beijing 100084, China(e-mail: likq@tsinghua.edu.cn).}%
}

% --- 页眉设置 ---
\markboth{IEEE TRANSACTIONS ON IMAGE PROCESSING,~Vol.~XX, No.~X, 2026}%
{Chen \MakeLowercase{\textit{et al.}}: Continuous Splatting meets Retinex}

\maketitle

% --- 摘要部分 ---
\begin{abstract}
Low-light image enhancement aims to recover clear images from low-illumination observations and is crucial for high-level downstream vision tasks. However, existing methods frequently encounter color distortion and structural artifacts when balancing global smooth illumination adjustment and local high-frequency detail recovery. To address these issues, we propose CGS-Retinex as the first low-light image enhancement framework based on explicit-implicit joint modeling. Our framework deeply integrates continuous Gaussian splatting with Retinex theory. Specifically, we represent the image grid as a continuous parameter field and propose a continuous Gaussian renderer to estimate the spatially continuous global illumination distribution. This approach fundamentally eliminates grid artifacts caused by discrete Gaussian sampling. Furthermore, we introduce an implicit neural representation to model reflectance independently. We leverage shallow high-frequency features to guide the network in accurately reconstructing degraded texture details. Within the Retinex framework, we incorporate physics-inspired brightness consistency constraints and illumination smoothness regularization to enable explicit illumination and implicit reflectance to maintain proper exposure and achieve high-fidelity recovery of high-frequency structures and colors. Extensive experiments demonstrate that CGS-Retinex significantly suppresses dark-region noise and overexposure while achieving exceptional high-frequency structural fidelity and color restoration by precisely decoupling illumination and texture. This work establishes a novel continuous physical representation paradigm for low-light image enhancement.
\end{abstract}

% --- 关键词部分 ---

% --- 第一章：Introduction ---
\section{Introduction}
\IEEEPARstart{L}{ow-light} image enhancement (LLIE) is a fundamental and challenging task in computer vision, aiming to restore visual quality from degraded observations with low illumination, low contrast, and complex noise. This task is crucial for downstream applications such as autonomous driving and all-weather surveillance \cite{ref1, ref2, ref3}. Over the past decades, LLIE techniques have shifted from traditional histogram equalization and Retinex-based models \cite{ref4, ref5} toward the deep learning paradigm. The widespread application of convolutional neural networks (CNNs) \cite{ref6, ref7, ref8, ref9, ref10, ref11, ref12}, Transformers, and recently emerging diffusion models \cite{ref13, ref14, ref15} has significantly advanced the performance of low-light image enhancement. Specifically, physics-inspired Retinex deep decoupling architectures \cite{ref16, ref17, ref18, ref19, ref20, ref21} provide strong physical interpretability by decomposing an image into illumination and reflectance maps.

Despite their significant progress, deep Retinex methods often struggle to balance smooth global illumination adjustment and local high-frequency detail recovery in complex real-world scenes. Furthermore, traditional pixel-level discrete modeling approaches are highly prone to producing color distortion, local overexposure, and severe structural artifacts under drastic brightness variations \cite{ref9, ref22}. To overcome the limitations of discrete representations, implicit neural representations (INRs) \cite{ref23} have been introduced to the LLIE task \cite{ref24, ref25}, capturing the intrinsic image structure through coordinate-driven continuous mappings. However, INRs suffer from significant spectral bias when modeling highly nonlinear high-frequency textures and lack explicit geometric control, making them inadequate for recovering severely degraded low-light details.

Concurrently, explicit radiance field modeling techniques represented by 3D Gaussian Splatting (3DGS) \cite{ref26} have revolutionized 3D reconstruction and rendering \cite{ref27, ref28, ref29, ref30}. This paradigm rapidly expanded to the 2D image domain and evolved into explicit image representation techniques including 2DGS \cite{ref31} and GaussianImage \cite{ref32, ref33, ref34}. By modeling images as numerous discrete Gaussian primitives, these methods demonstrate exceptional representation efficiency in image compression, super-resolution \cite{ref35}, and dataset distillation \cite{ref36}. Recent studies have explored Gaussian splatting for low-light image enhancement \cite{ref37, ref38}. However, most existing 2DGS schemes rely on a discrete splatting mechanism that generates pixels by querying nearest-neighbor primitives or aggregating discrete points. Such discreteness introduces discontinuous grid artifacts into the continuous illumination field, severely compromising the physical prior of illumination smoothness essential to Retinex theory.

To resolve the aforementioned trade-offs, we propose CGS-Retinex as the first low-light image enhancement framework based on explicit-implicit joint modeling. We deeply integrate Continuous Gaussian Splatting (CGS) with implicit neural representation (INR) within the Retinex architecture to establish a novel continuous physical representation paradigm. Specifically, we propose a continuous Gaussian renderer for the illumination field. Unlike traditional discrete primitive sampling, we model Gaussian attributes as a spatially continuous parameter field. By interpolating and blending the Gaussian parameter field over continuous coordinates, we achieve a spatially smoother and more continuous illumination estimation. This approach significantly suppresses the grid discontinuities and artifacts commonly caused by discrete splatting in dense prediction tasks. To model the reflectance, we employ a coordinate-based INR and introduce shallow high-frequency features via skip connections to guide the network in precisely reconstructing degraded edges and textures within a continuous coordinate space. The primary contributions of this work are summarized as follows:

\begin{enumerate}
\item We propose CGS-Retinex as the first low-light image enhancement framework based on explicit-implicit joint modeling. By deeply integrating the explicit modeling of 2D Gaussian splatting with the continuous mapping of implicit neural representations under strict Retinex physical constraints, our framework establishes a novel physical representation paradigm to decouple complex illumination adjustment from high-frequency detail recovery.
\item To overcome the grid artifacts frequently caused by the discrete sampling of 2DGS primitives in dense prediction tasks, we propose an artifact-free continuous Gaussian rendering mechanism. By lifting the input grid into a spatially continuous parameter field and directly performing bilinear interpolation and distance-based rendering within the continuous domain, this mechanism mathematically guarantees an exceptionally smooth global illumination distribution.
\item We introduce a feature-guided implicit reflectance decoding network that leverages a coordinate network to independently model reflectance residuals. By integrating shallow high-frequency features via skip connections, this architecture achieves precise reconstruction of degraded textures and high-fidelity color recovery within a continuous coordinate space, thereby significantly enhancing noise suppression in dark regions.
\end{enumerate}

% --- 第二章：Related Work ---
\section{Related Work}
Our proposed CGS-Retinex lies at the intersection of explicit-implicit joint modeling and low-level vision tasks, with a particular focus on deeply integrating continuous Gaussian splatting and implicit reflectance modeling for low-light image enhancement. Therefore, we divide the related literature into two main categories. First, we review recent advancements in low-light image enhancement. Second, we summarize the overall development of Gaussian splatting and explicit image representations.

\subsection{Low-Light Image Enhancement}
Low-light image enhancement (LLIE) aims to restore high-quality visual content from degraded observations characterized by suboptimal illumination, low contrast, and complex noise \cite{ref1, ref2}. To address this highly ill-posed inverse problem, early studies extensively explored Retinex physical priors. For instance, LIME \cite{ref4} precisely estimates the illumination map using structural priors, and ChebyLighter \cite{ref5} introduces optimal Chebyshev approximation for curve estimation. Subsequently, Retinex-Net \cite{ref20} pioneered the deep learning paradigm of decomposing an image into illumination and reflectance layers. Recently, convolutional neural networks (CNNs) have greatly advanced this field. Various multi-scale and residual architectures, such as FMR-Net \cite{ref7}, the re-parameterized residual network FRR-NET \cite{ref8}, the enhanced fusion iterative network EFINet \cite{ref9}, and R2RNet \cite{ref12} for mapping real low-light observations to normal-light counterparts, significantly improve the signal-to-noise ratio and color fidelity of images.

To alleviate the heavy reliance on expensive paired data, unsupervised and zero-shot learning have emerged as mainstream approaches. Zero-DCE \cite{ref39} and its successor Zero-DCE++ \cite{ref40} pioneered zero-reference deep curve estimation to transform image enhancement into iterative image-specific curve mapping. EnlightenGAN \cite{ref41} achieves deep illumination adjustment by leveraging unpaired data and attention mechanisms. Similarly, ZERO-IG \cite{ref19} explores zero-shot illumination-guided joint denoising. Furthermore, Fu et al. \cite{ref10} and Zhang et al. \cite{ref22} achieve superior denoising and enhancement under extreme conditions without task-related data through simple paired instances and noise autoregressive paradigms.

The evolution of network architectures places unprecedented emphasis on lightweight design and physical interpretability. Bai et al. \cite{ref6} and Chen et al. \cite{ref11} designed ultra-lightweight real-time enhancement architectures for mobile devices and embedded automotive vision systems, respectively, while Ma et al. \cite{ref42} proposed a calibration network for fast, flexible, and robust enhancement. Concurrently, by deeply unfolding traditional optimization algorithms, methods such as RUAS \cite{ref16}, URetinex-Net \cite{ref18}, and adaptive unfolding total variation networks \cite{ref17} successfully integrate the rigor of physical models with the feature representation capabilities of neural networks.

Recently, MambaLLIE \cite{ref21} has integrated global-local state space models into implicit Retinex perception, empowering the effective modeling of long-range illumination dependencies. Concurrently, diffusion models have demonstrated significant potential in addressing complex degradations by leveraging disruptive generative priors. Specifically, Retinex-driven latent diffusion models such as Lightendiffusion \cite{ref13}, the training-free Aglldiff \cite{ref14}, and zero-shot latent diffusion models \cite{ref15} have successively broken through the performance ceilings of generative low-light enhancement.

Despite the significant progress of discrete pixel-grid methods, their fixed spatial resolutions frequently induce jagged structural artifacts in extremely underexposed regions. To overcome these limitations, implicit neural representations (INRs) \cite{ref23} have gained attention for modeling images as continuous functional mappings. By taking spatial coordinates as inputs to multilayer perceptrons, INRs enable resolution-independent image representation. Yang et al. \cite{ref24} first proposed an INR-based model for synergistic low-light enhancement by integrating continuous coordinate systems with local illumination features. Chobola et al. \cite{ref25} developed context-based fast neural implicit representations to accelerate the enhancement process. However, pure implicit networks suffer from inherent spectral bias when fitting highly nonlinear high-frequency signals. Moreover, the absence of explicit geometric constraints on local textures leads to color distortion and over-smoothing during the recovery of severely degraded details. This bottleneck prompts us to investigate whether introducing explicit modeling while retaining the advantages of continuous coordinate mapping can overcome these shortcomings.

Accordingly, we incorporate a feature-guided implicit reflectance decoder into CGS-Retinex. By fusing coordinate-based continuous modeling with shallow high-frequency features, our network accurately reconstructs degraded edges within a continuous coordinate space. This design effectively addresses the inadequacy of pure implicit representations in recovering complex reflectance textures.

\subsection{Gaussian Splatting and Explicit Image Representation}
Novel view synthesis has undergone a profound paradigm evolution since Neural Radiance Fields (NeRF) \cite{ref43} pioneered continuous implicit scene modeling via volume rendering. Recently, 3D Gaussian Splatting (3DGS) \cite{ref26} has broken the rendering speed bottleneck of implicit representations by leveraging explicit parametric representations based on 3D Gaussian primitives and fast tile-based differentiable rasterization. Serving as a cornerstone for large-scale scene reconstruction, 3DGS offers high flexibility due to its explicit nature and achieves remarkable success in dynamic urban scenes and autonomous driving. Street Gaussians \cite{ref27}, DrivingGaussian \cite{ref29}, CityGaussian \cite{ref30}, and Momentum-GS \cite{ref44} achieve high-fidelity rendering of large-scale urban scenes by integrating dynamic tracking with spatio-temporal decoupling. GaussianPro \cite{ref45} significantly improves surface reconstruction precision through a progressive propagation strategy. Similarly, Speedy-splat \cite{ref46} and Mvsgaussian \cite{ref28} substantially enhance reconstruction generalization and efficiency by leveraging sparse primitive acceleration and multi-view stereo priors. To model physical phenomena and large-span spatio-temporal motions, Projectile Motion Gaussian (PMGS) \cite{ref47}, Physics-Event enhanced Gaussian (PEGS) \cite{ref48}, and Periodic Vibration Gaussian \cite{ref49} successfully incorporate strict physical laws into the Gaussian parameter field.

Moreover, leveraging the powerful generative capabilities of 2D and 3D diffusion models, GaussianDreamer \cite{ref50}, DriveDreamer4D \cite{ref51}, and ReconDreamer \cite{ref52} enable rapid 3D Gaussian generation from text or single images and 4D driving scene recovery. For exploring adverse environments, initial studies have integrated 3DGS into low-light 3D reconstruction. Specifically, LL-Gaussian \cite{ref53} and LLGS \cite{ref54} pioneer joint image enhancement and novel view synthesis in low-light and pitch-dark scenarios. However, these methods rely heavily on multi-view geometric constraints, making them difficult to adapt directly for single-view enhancement.

Driven by the success of 3DGS, recent studies have reduced the dimensionality of primitives to achieve pure 2D image representation. Specifically, 2DGS \cite{ref31} compresses 3D ellipsoids into flattened 2D splats, thereby eliminating volumetric artifacts from 3D rendering and offering a fresh perspective on image modeling. Subsequently, GaussianImage \cite{ref34} innovatively employs 2D Gaussians for single-image representation, establishing a novel paradigm that represents images through Gaussian splatting primitives. This framework achieves a rendering rate of 1000 FPS and competitive image compression ratios. Building upon these advancements, Instant GaussianImage \cite{ref33} and LIG \cite{ref32} further compress gigapixel-level images by employing adaptive primitive splitting and multi-level splatting. Currently, 2DGS-based image representations have moved beyond simple fitting and are extensively applied to various low-level and high-level vision tasks. Notable applications include arbitrary-scale super-resolution (GaussianSR \cite{ref35}), sparse dataset distillation beyond pixel grids \cite{ref36}, and vision-language model alignment based on compressed Gaussian representations \cite{ref55}.

Given the superior representation efficiency of GaussianImage, recent studies have extended its application to single low-light image enhancement. Specifically, LL-GaussianImage \cite{ref37} introduces an efficient parameterized image representation to achieve zero-shot enhancement via 2D Gaussian splatting. Meanwhile, LL-GaussianMap \cite{ref38} leverages 2D Gaussian networks to generate spatially variant gain maps, enabling adaptive brightness adjustment for input images. While these pioneering works validate the significant potential of 2D explicit Gaussians in LLIE, they still rely on the original discrete splatting mechanism, which renders pixels by aggregating discrete point primitives or querying nearest-neighbor Gaussian attributes. In Retinex decomposition, global illumination must strictly adhere to the physical prior of spatial smoothness. However, this discrete sampling and aggregation mechanism inevitably introduces sharp transitions and grid artifacts into the continuous illumination field.

To fundamentally resolve this contradiction, we depart from the traditional discrete splatting paradigm and propose a Continuous Gaussian Splatting (CGS) renderer. Our approach lifts the input grid into a spatially continuous Gaussian parameter field, enabling bilinear interpolation and distance-based radiance rendering directly within the continuous domain. Consequently, this mechanism mathematically guarantees an infinitely smooth transition of attributes at any coordinate. This innovation completely eliminates the grid noise caused by discrete primitives while achieving a seamless physical alignment between Gaussian explicit representation and Retinex illumination smoothing theory. By unifying these two components, our approach establishes a novel continuous physical representation paradigm for low-light image enhancement.

% --- 第三章：Proposed Method ---
\section{Proposed Method}
\begin{figure*}[!t]
\centering
\includegraphics[width=\linewidth]{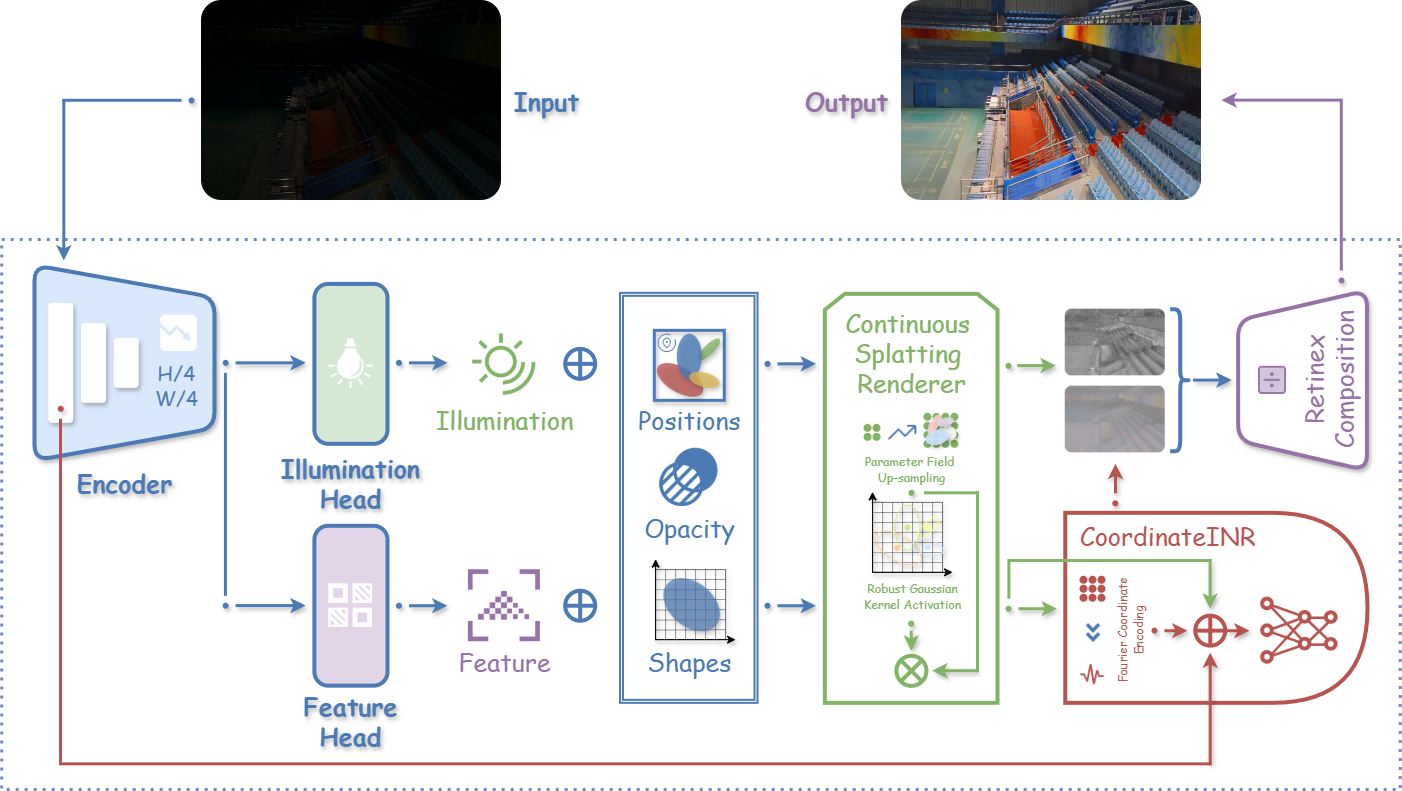}
\caption{Overall architecture of the CGS-Retinex framework. Given a low-light input image, the encoder extracts multi-level features, routing them into parallel illumination and feature branches. Both branches leverage the Continuous Splatting Renderer to respectively generate a spatially continuous illumination distribution and a context feature field. During implicit reconstruction, the CoordinateINR integrates this feature field alongside shallow high-frequency features to accurately predict the reflectance residual. Finally, the Retinex Composition module combines the continuous illumination map and the reflectance residual, deriving the high-fidelity enhanced image under strict physical constraints.}
\label{fig1}
\end{figure*}

In this section, we detail CGS-Retinex, the proposed low-light image enhancement framework based on continuous Gaussian splatting and implicit reflectance modeling. First, as illustrated in Fig. \ref{fig1}, we provide a comprehensive overview of the architecture and introduce our novel Retinex-based decoupling paradigm. Subsequently, we provide an in-depth analysis of the continuous Gaussian parameter prediction head, specifically comparing the design principles of the illumination and feature heads. Next, as our core theoretical innovation, we derive and construct the Continuous Splatting Renderer incorporating a learnable temperature coefficient. We then introduce a coordinate-driven implicit neural representation to recover high-frequency textures. Finally, we detail a systematically constructed joint loss function system inspired by physical laws.

\subsection{Overall Architecture and Continuous Physical Retinex Decoupling Paradigm}
Classical Retinex theory assumes that an observed image $I \in \mathbb{R}^{3 \times H \times W}$ can be decomposed into the pixel-wise product of an illumination map $\mathcal{L}$ and a reflectance map $\mathcal{R}$. However, under extremely dark or non-uniform lighting conditions, traditional discrete grid assumptions often induce blocky artifacts in illumination estimation, while the reflectance suffers from severe color degradation and noise amplification. To fundamentally overcome these limitations, we depart from the traditional perspective of treating images as discrete pixel matrices and establish a Retinex physical representation based on spatially continuous fields.

Specifically, given a low-light input image $I_{in}$, the CGS-Retinex framework first employs a hierarchical feature encoder $\mathcal{E}$ incorporating dense residual connections to extract deep contextual features $\mathcal{F}_{enc}$ and shallow skip features $\mathcal{F}_{skip}$ containing high-frequency spatial details:
\begin{equation}
\mathcal{F}_{skip}, \mathcal{F}_{enc} = \mathcal{E}(I_{in})
\end{equation}

Departing from conventional end-to-end mapping mechanisms that directly regress $\mathcal{L}$ and $\mathcal{R}$, we route the deep features $\mathcal{F}_{enc}$ into two parallel continuous field parameterization branches: an illumination prediction head $\mathcal{H}_{illu}$ and a feature prediction head $\mathcal{H}_{feat}$. These heads generate continuous Gaussian primitive parameter sets, denoted as $\Theta_{illu}$ and $\Theta_{feat}$, to represent distinct physical properties. Subsequently, the proposed continuous Gaussian renderer $\mathcal{R}_{cgs}$ integrates these discrete parameter primitives to formulate continuous spatial field distributions. We formalize the generation of the illumination field $\mathcal{L}_{cgs}$ as follows:
\begin{equation}
\mathcal{L}_{cgs}(p)=\mathcal{R}_{cgs}(p \mid \Theta_{illu}), \quad \forall p \in \Omega
\end{equation}
where $p=(x, y)$ denotes an arbitrary query coordinate within the continuous space $\Omega$. To prevent numerical instability and color bleaching caused by division operations in extremely low-illumination regions, we apply an adaptive truncation mechanism to the illumination map, physically constraining its lower bound to $\tau=0.05$:
\begin{equation}
\mathcal{L}=\max(\mathcal{L}_{cgs}, \tau)
\end{equation}

Upon obtaining the globally smooth illumination $\mathcal{L}$, we directly derive the coarse base reflectance $\mathcal{R}_{base}$ through the classical Retinex inversion:
\begin{equation}
\mathcal{R}_{base} = I_{in} / \mathcal{L}
\end{equation}
Unlike existing Retinex-based methods that typically adopt $\mathcal{R}_{base}$ as the final output or employ complex convolutional networks to map the final reflectance $\mathcal{R}$, we introduce a joint explicit-implicit modeling paradigm to refine the physical Retinex decomposition and effectively decouple noise from textures. Specifically, we model the high-frequency texture residual $\mathcal{R}_{res}$ by leveraging a coordinate-driven implicit neural representation $\Phi_{INR}$ conditioned on continuous context features $C_{feat}$ from the renderer $\mathcal{R}_{cgs}$ and shallow skip features $\mathcal{F}_{skip}$:
\begin{equation}
\mathcal{R}_{res}(p) = \Phi_{INR}(\gamma(p), C_{feat}(p), \mathcal{F}_{skip}(p))
\end{equation}

We formulate the final enhanced image $I_{out}$ by linearly combining the physical reflectance base and the implicit high-frequency residual. To bound the values within the valid color space, we apply a clamping function:
\begin{equation}
I_{out} = \text{Clamp}(\mathcal{R}_{base} + \mathcal{R}_{res}, 0, 1)
\end{equation}

\begin{figure}[!t]
\centering
\includegraphics[width=\linewidth]{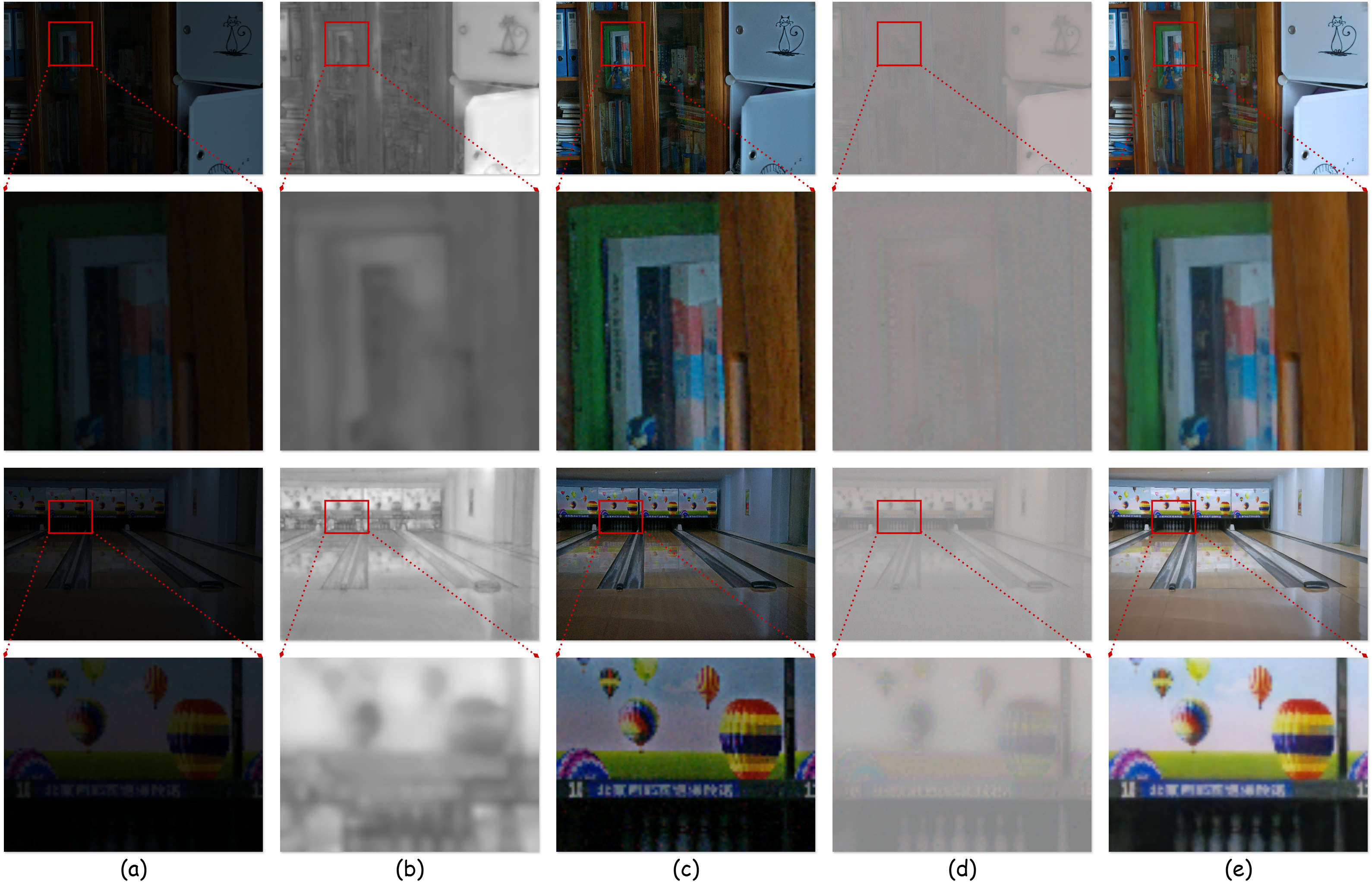}
\caption{Intermediate feature decoupling and local detail visualization of CGS-Retinex. (a) Original low-light input; (b) estimated spatial illumination map $\mathcal{L}$; (c) base reflectance map $\mathcal{R}_{base}$; (d) residual map $\mathcal{R}_{res}$ generated by $\Phi_{INR}$; (e) final enhanced output.}
\label{fig2}
\end{figure}

To verify the physical interpretability of our network, Fig. \ref{fig2} illustrates the feature decoupling process and local details during forward propagation. Given a low-light input (a), the model first estimates the illumination map (b) and derives the base reflectance map (c). Observing the zoomed-in regions reveals that while (c) successfully restores the global color, it suffers from high-frequency texture blurring due to the extremely low signal-to-noise ratio. Consequently, we introduce the INR module to specifically predict the detail compensation residual (d), precisely extracting the missing structures and edges. The final fused output (e) achieves both high color fidelity and superior clarity, directly validating the effectiveness and mathematical rigor of our strategy.

\subsection{Attribute-Decoupled Continuous Gaussian Parameter Prediction Heads}
To map the input features into the spatial geometry and attribute parameters required for continuous Gaussian splatting, we design the Gaussian Parameter Head. Within the CGS-Retinex framework, illumination and texture exhibit fundamentally distinct frequency characteristics. While global illumination manifests as low-frequency gradual transitions across the spatial domain, texture features contain abundant high-frequency local variations. Consequently, we separately instantiate the illumination prediction head $\mathcal{H}_{illu}$ and the feature prediction head $\mathcal{H}_{feat}$ to accommodate these distinct physical properties.

For the input feature tensor $\mathcal{F}_{enc} \in \mathbb{R}^{C \times H_d \times W_d}$, the prediction head first extracts base features via parameter-sharing nonlinear convolutions and subsequently partitions them into geometric and attribute branches. For each grid center $(i, j)$, the network estimates a corresponding Gaussian primitive $G_{i, j}$. First, we establish a regular base grid $X_{base}$ within the normalized image coordinate system $[-1, 1]^2$. The prediction network then regresses a minor spatial offset $\Delta_{i, j}$ to derive the continuous center coordinate $\mu_{i, j}$ of each Gaussian primitive:
\begin{equation}
\mu_{i, j} = X_{base}(i, j) + \frac{2}{\min(H, W)} \tanh(\Delta_{i, j})
\end{equation}

The covariance matrix $\Sigma$ of the Gaussian kernel determines the influence range and anisotropic shape of the Gaussian primitive. To guarantee positive definiteness and facilitate the computation of the Mahalanobis distance, we directly predict the inverse covariance matrix $\Sigma^{-1}$. We construct this inverse matrix using the predicted anisotropic scaling factors $S = [s_x, s_y]^T$ and the rotation angle $\theta$. To prevent numerical singularities, we map the scaling factors through a Sigmoid activation function $\sigma$:
\begin{equation}
S = \frac{2}{\min(H, W)} \sigma(\Delta S) + \epsilon, \quad \theta = \pi \tanh(\Delta \theta)
\end{equation}

By multiplying the rotation matrix $\mathcal{R}(\theta)$ and the diagonal scaling matrix $D(S)$, we construct the geometric transformation matrix and subsequently derive the specific components $a, b,$ and $c$ of the inverse covariance matrix:
\begin{equation}
\Sigma^{-1} = (\mathcal{R}(\theta) D(S) D(S)^T \mathcal{R}(\theta)^T)^{-1} = \begin{bmatrix} a & b \\ b & c \end{bmatrix}
\end{equation}

We precisely calculate these matrix elements using the following expressions:
\begin{equation}
\left\{
\begin{aligned}
a &= \left(\frac{\cos \theta}{s_x}\right)^2 + \left(\frac{\sin \theta}{s_y}\right)^2 \\
b &= -\sin \theta \cos \theta \left(\frac{1}{s_x^2} - \frac{1}{s_y^2}\right) \\
c &= \left(\frac{\sin \theta}{s_x}\right)^2 + \left(\frac{\cos \theta}{s_y}\right)^2
\end{aligned}
\right.
\end{equation}

Furthermore, the network predicts an opacity factor $\alpha \in (0, 1)$ and a feature attribute $\mathcal{A}$ for each Gaussian primitive. Within the illumination head $\mathcal{H}_{illu}$, the attribute $\mathcal{A}_{illu} \in \mathbb{R}^1$ represents the single-channel scalar illumination intensity, where we apply a Softplus activation function to guarantee strict non-negativity. Correspondingly, within the feature head $\mathcal{H}_{feat}$, the attribute $\mathcal{A}_{feat} \in \mathbb{R}^{64}$ denotes a high-dimensional semantic vector. Unlike traditional 3D Gaussian Splatting, our prediction head implicitly binds image priors to the Gaussian parameters. Consequently, this architectural design enables efficient feed-forward inference and generalizes seamlessly to input images of arbitrary resolutions.

\subsection{Continuous Gaussian Renderer with Learnable Temperature}
Originally developed for novel view synthesis, traditional 3D Gaussian Splatting relies on a point cloud rasterization process that lacks spatial smoothing constraints between adjacent pixels. To address this fundamental limitation, we propose the continuous Gaussian splatting renderer $\mathcal{R}_{cgs}$. By integrating neighborhood multi-Gaussian blending with an adaptive temperature alpha composition, this renderer achieves perfectly smooth interpolation at arbitrary spatial coordinates.

To evaluate a given continuous query coordinate $p=(x_q, y_q)$, we transcend the limitation of querying a single nearest-neighbor Gaussian primitive. Instead, we project $p$ back into the feature space and deploy an unfold mechanism to extract a local $K \times K$ set of neighborhood Gaussians $\mathcal{N}(p)$, where we empirically set $K=3$. For the $k$-th Gaussian primitive $G_k=\{\mu_k, \Sigma_k^{-1}, \alpha_k, \mathcal{A}_k\}$ within $\mathcal{N}(p)$, we first formulate the spatial offset vector $d_k=[dx_k, dy_k]^T$ between the query point $p$ and the Gaussian center $\mu_k$:
\begin{equation}
d_k=p-\mu_k
\end{equation}
Subsequently, we compute the corresponding squared Mahalanobis distance $D_k$:
\begin{equation}
D_k=d_k^T\Sigma_k^{-1}d_k=a_k dx_k^2+2b_k dx_k dy_k+c_k dy_k^2
\end{equation}

To prevent gradient explosion and control the Gaussian tails, we truncate the distance term at a maximum threshold of $\lambda=10$, formulating the exponential decay weight of the primitive as follows:
\begin{equation}
G_k(p)=\exp\left(-\frac{1}{2}\min(D_k, \lambda)\right)
\end{equation}

Traditional alpha compositing in 2D grids frequently induces depth discontinuities, yielding hard boundaries within the rendered illumination maps. To address this critical flaw, we significantly restructure the alpha compositing process into a normalized weighting mechanism guided by an opacity prior. Specifically, we define the raw blending weight as the product of the Gaussian decay and the opacity factor:
\begin{equation}
w_{raw, k}(p)=G_k(p)\cdot\alpha_k
\end{equation}

To adaptively balance extreme smoothness and sharpness preservation, we incorporate a learnable temperature scalar $T$. We define this temperature using a Softplus function with a minimal offset to ensure strict positivity and prevent division-by-zero errors during optimization:
\begin{equation}
T_{adp} = \ln(1 + \exp(T)) + 10^{-4}
\end{equation}

Leveraging this adaptive temperature, we perform Softmax normalization on the blending weights within the local $K \times K$ neighborhood $\mathcal{N}(p)$:
\begin{equation}
w_{norm, k}(p) = \frac{\exp(w_{raw, k}(p) / T_{adp})}{\sum_{j \in \mathcal{N}(p)} \exp(w_{raw, j}(p) / T_{adp})}
\end{equation}

Finally, we compute the rendered attribute value $O(p)$ at the query coordinate $p$ by integrating the weighted attributes of all neighborhood primitives:
\begin{equation}
O(p)=\sum_{k \in \mathcal{N}(p)} w_{norm, k}(p) \cdot \mathcal{A}_k
\end{equation}
This mechanism fundamentally eliminates the aliasing artifacts inherent to discrete grid interpolation within high-dimensional feature spaces, empowering the network to generate exceptionally smooth and physically coherent illumination fields. As $T_{adp}$ approaches a large value, the illumination converges toward a uniform distribution. Conversely, through iterative backpropagation, the network autonomously discovers the optimal $T_{adp}$ to preserve essential gradient variations at structural boundaries.

\subsection{Coordinate-Driven Implicit Neural Reflectance Representation}
Traditional convolutional neural networks frequently over-smooth high-frequency noise within extremely dark regions of low-light images, inevitably causing severe texture loss. To reconstruct high-frequency reflectance details with nearly infinite resolution and superior fidelity, we integrate a coordinate-driven implicit neural representation network $\Phi_{INR}$ into the residual path. Utilizing a multi-layer perceptron (MLP) as the primary backbone, this network requires specialized coordinate encoding. To enable the shallow MLP layers to effectively perceive high-frequency spatial variations, we first map the normalized 2D spatial coordinates $p$ into a high-dimensional frequency space using a fixed random Fourier basis $\mathcal{B} \in \mathbb{R}^{2 \times \frac{D_{hid}}{2}}$:
\begin{equation}
\left\{
\begin{aligned}
p_{proj} &= p^T \mathcal{B} \\
\gamma(p) &= [\sin(p_{proj}), \cos(p_{proj})]
\end{aligned}
\right.
\end{equation}

Subsequently, we concatenate the high-frequency coordinate encoding $\gamma(p)$, the spatially smooth context features $C_{feat}(p)$ from the Gaussian renderer, and the resolution-aligned shallow skip features $\mathcal{F}_{skip}(p)$ along the channel dimension. This composite tensor serves as the primary input for the implicit network:
\begin{equation}
\Gamma_{in} = \text{Concat}(\gamma(p), C_{feat}(p), \mathcal{F}_{skip}(p))
\end{equation}

Unlike conventional INRs that rely solely on coordinate mapping and frequently lack awareness of global semantics or illumination distributions, our $\Phi_{INR}$ architecture deeply integrates $C_{feat}$ derived from the continuous Gaussian field. This integration endows the INR with essential spatial geometry and illumination priors, effectively guiding the reconstruction of high-frequency details within a physically consistent context.

To perform joint nonlinear decoding of coordinates and high-dimensional features, we reconstruct the texture residual using a four-layer cascaded $1 \times 1$ convolutional network. To maintain a concise mathematical representation, we denote the mapping of the first three hidden layers as a composite nonlinear function $\mathcal{F}_{mlp}$. Following the hierarchical feature extraction from the input tensor $\Gamma_{in}(p)$, the final layer employs a tanh activation to strictly bound the residual output within the interval $[-1, 1]$. The complete decoding mapping is formulated as follows: The composite mapping $\mathcal{F}_{mlp}(\cdot) = (\phi_3 \circ \phi_2 \circ \phi_1)(\cdot)$ encapsulates the nonlinear refinement process of deep features. To ensure the statistical stability of the spatial features, we construct the initial hidden layer $\phi_1$ using a convolution operation coupled with Group Normalization and GELU activation:
\begin{equation*}
\phi_1(X) = \text{GELU}(\text{GroupNorm}(W_1 * X + b_1))
\end{equation*}
Subsequent refinement layers $\phi_2$ and $\phi_3$ bypass normalization, instead adopting a standard joint operator composed of a $1 \times 1$ convolution and a GELU activation function.

Notably, we intentionally avoid downscaling the magnitude of the residual $\mathcal{R}_{res}$ in the final prediction. This architectural choice liberates the INR from the conservative constraints of prior methods, empowering it to autonomously govern the compensation amplitude for both color saturation and high-frequency textures. Consequently, this unconstrained strategy profoundly unlocks the network's capability to reconstruct high-purity colors and razor-sharp edges within severely degraded dark areas.

\subsection{Joint Loss Function System Under Physical Constraints}
To simultaneously optimize the spatial smoothness of the continuous Gaussian rendering, the physical validity of the Retinex decoupling, and the perceptual quality of the final image, we formulate a comprehensive joint loss function encompassing six distinct sub-terms. This holistic system strictly governs the network behavior throughout the optimization process. For the subsequent mathematical definitions, let $I_{gt}$, $I_{out}$, and $\mathcal{L}$ denote the ground-truth clear image, the final predicted output, and the estimated illumination map, respectively.

First, we employ the Charbonnier loss as a fundamental pixel-level constraint to significantly elevate the Peak Signal-to-Noise Ratio (PSNR) and compel the network to focus on challenging high-frequency details. Compared to the standard $L_2$ penalty, the Charbonnier loss effectively mitigates edge blurring:
\begin{equation}
L_{rec}=\frac{1}{N}\sum_{i=1}^{N}\sqrt{(I_{out}(i)-I_{gt}(i))^2+\epsilon^2}
\end{equation}
where $\epsilon=10^{-3}$ represents a minimal regularization constant introduced to prevent gradient singularities at the origin.

To preserve the fidelity of spatial structures and local contrast, we incorporate a structural similarity (SSIM) loss constraint:
\begin{equation}
\begin{split}
L_{ssim} &= 1 - \text{SSIM}(I_{out}, I_{gt}) \\
&= 1 - \frac{2\mu_x\mu_y+C_1}{\mu_x^2+\mu_y^2+C_1} \cdot \frac{2\sigma_{xy}+C_2}{\sigma_x^2+\sigma_y^2+C_2}
\end{split}
\end{equation}

Furthermore, we introduce a perceptual feature loss to align the enhanced results with the human visual system's perception. By leveraging a pre-trained VGG-16 network, this loss computes the mean squared error between deep feature representations. To ensure consistent feature scales, we perform instance normalization on the images using a predefined mean $\mu_{vgg}$ and variance $\sigma_{vgg}$ prior to feature extraction:
\begin{equation}
\left\{
\begin{aligned}
I_{norm} &= \frac{I - \mu_{vgg}}{\sigma_{vgg}} \\
L_{vgg} &= \frac{1}{M_j}\|\Phi_{vgg}(I_{out})_j-\Phi_{vgg}(I_{gt})_j\|_2^2
\end{aligned}
\right.
\end{equation}

According to the physical priors of the Retinex theory, natural illumination distributions inherently exhibit piecewise smoothness. To suppress potential local abruptions and high-frequency noise interference, we impose an anisotropic Total Variation (TV) loss on the illumination map $\mathcal{L}$ generated by the continuous Gaussian prediction head:
\begin{equation}
\left\{
\begin{aligned}
\nabla_x\mathcal{L} &= \mathcal{L}_{:,:,1:,:} - \mathcal{L}_{:,:,:-1,:} \\
\nabla_y\mathcal{L} &= \mathcal{L}_{:,:,:,1:} - \mathcal{L}_{:,:,:,:-1} \\
L_{tv} &= 2C_h(\nabla_x\mathcal{L})^2 + 2C_w(\nabla_y\mathcal{L})^2
\end{aligned}
\right.
\end{equation}
where $C_w$ and $C_h$ represent the spatial normalization coefficients. Furthermore, to counteract the severe color shifts that frequently occur during extreme low-light enhancement, we introduce a color consistency loss. By calculating the cosine similarity across pixel channels, this constraint explicitly minimizes the angular distance between the predicted image and the ground truth within the color vector space:
\begin{equation}
L_{color}=1-\frac{1}{N}\sum_i\frac{\langle I_{out}(i), I_{gt}(i)\rangle}{\|I_{out}(i)\|_2\cdot\|I_{gt}(i)\|_2}
\end{equation}

Conventional approaches typically enforce a strict equivalence between the element-wise product of the reconstructed image $I_{out}$ and the illumination map $\mathcal{L}$ against the original input image $I_{in}$. However, in real-world physical processes, the camera Image Signal Processing (ISP) pipeline applies complex color compression. Enforcing the exact constraint $I_{out}\odot\mathcal{L}\approx I_{in}$ across all three RGB channels severely impedes the capacity of the network to autonomously restore high-purity and highly saturated colors. To circumvent this limitation, we innovatively propose a physical loss mechanism that imposes the reconstruction consistency constraint exclusively within the grayscale domain. Let $\mathcal{F}_{gray}(\cdot)$ denote the channel-wise mean pooling operation. We mathematically formulate the grayscale intensities $Y_{in}$ and $Y_{rec}$ as follows:
\begin{equation}
\left\{
\begin{aligned}
Y_{in} &= \mathcal{F}_{gray}(I_{in}) = \frac{1}{3}\sum_{c \in \{R, G, B\}} I_{in, c} \\
Y_{rec} &= \mathcal{F}_{gray}(I_{out}\odot\mathcal{L}) = \frac{1}{3}\sum_{c \in \{R, G, B\}} I_{out, c}\cdot\mathcal{L}
\end{aligned}
\right.
\end{equation}

Employing the Charbonnier penalty, we relax the rigid constraints on color restoration while strictly maintaining the brightness-based Retinex degradation consistency:
\begin{equation}
L_{phy}=\sqrt{(Y_{rec}-Y_{in})^2+\epsilon^2}
\end{equation}

Consequently, we formulate the comprehensive optimization objective as a weighted sum of the aforementioned terms:
\begin{equation}
\begin{aligned}
L_{total} = \,\, & \lambda_1 L_{rec} + \lambda_2 L_{ssim} + \lambda_3 L_{vgg} \\
& + \lambda_4 L_{tv} + \lambda_5 L_{phy} + \lambda_6 L_{color}
\end{aligned}
\end{equation}
Based on extensive empirical validation, we assign the following hyperparameter weights: $\lambda_1=1.0$, $\lambda_2=0.5$, $\lambda_3=0.05$, $\lambda_4=0.05$, $\lambda_5=0.05$, and $\lambda_6=0.05$. Synergizing with the dual advantages of continuous Gaussian splatting and implicit residual modeling, this meticulously designed loss system empowers the network to achieve exceptional photometric and structural consistency at a minimal computational cost.

\begin{figure*}[!t]
\centering
\includegraphics[width=\linewidth]{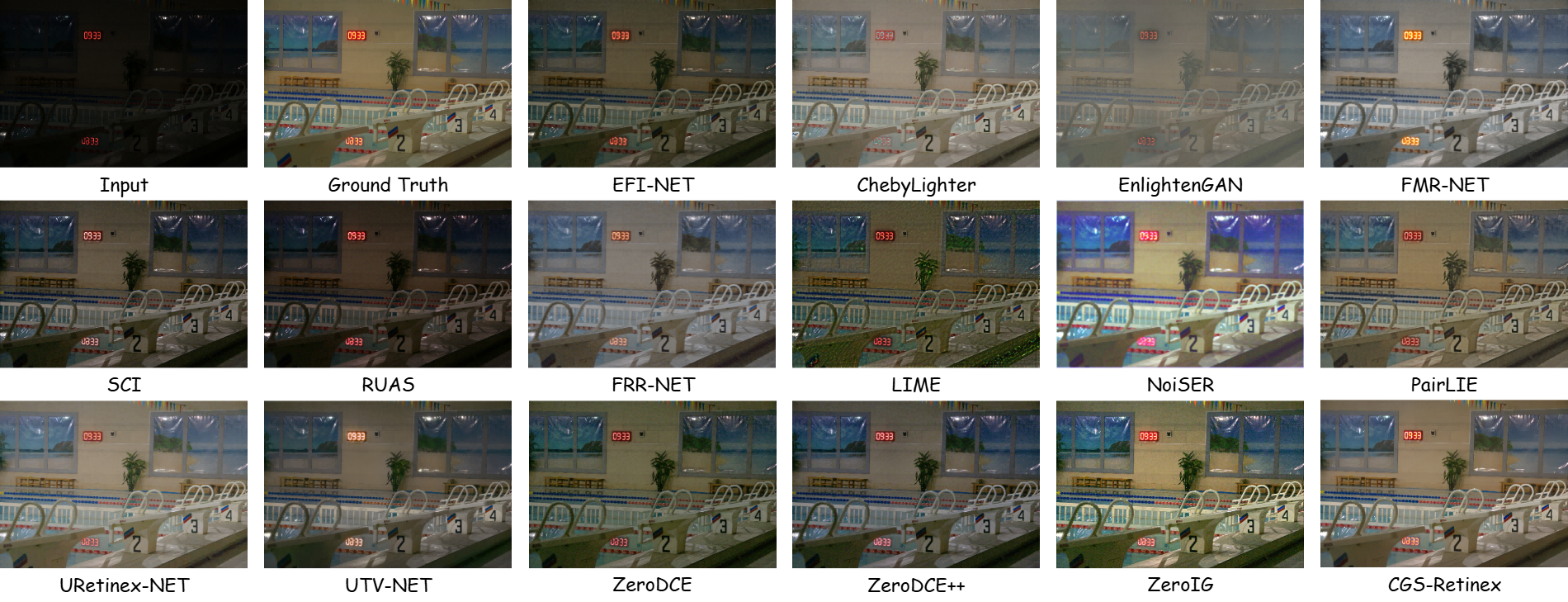}
\caption{Visual comparison of CGS-Retinex and state-of-the-art methods on the LOL dataset.}
\label{fig3}
\end{figure*}

\section{Experiments}
\subsection{Experimental Setup}
\textbf{Datasets.} To evaluate the effectiveness and generalization capabilities of our CGS-Retinex model, we conduct extensive experiments on two benchmark datasets: the LOL dataset \cite{ref20} and the LSRW dataset \cite{ref12}. We rigorously partition the data, allocating the samples into training and testing sets at a strict 9:1 ratio. As the pioneering publicly available paired dataset specifically constructed for supervised low-light image enhancement, the LOL dataset comprises synthetically generated low-light images alongside real-world normal-light observations. Furthermore, the LSRW dataset serves as the first large-scale paired image benchmark captured entirely in real-world scenarios, consisting of two independent subsets captured using a Huawei P40 Pro smartphone and a Nikon D7500 DSLR camera, respectively.

\begin{figure*}[!t]
\centering
\includegraphics[width=\linewidth]{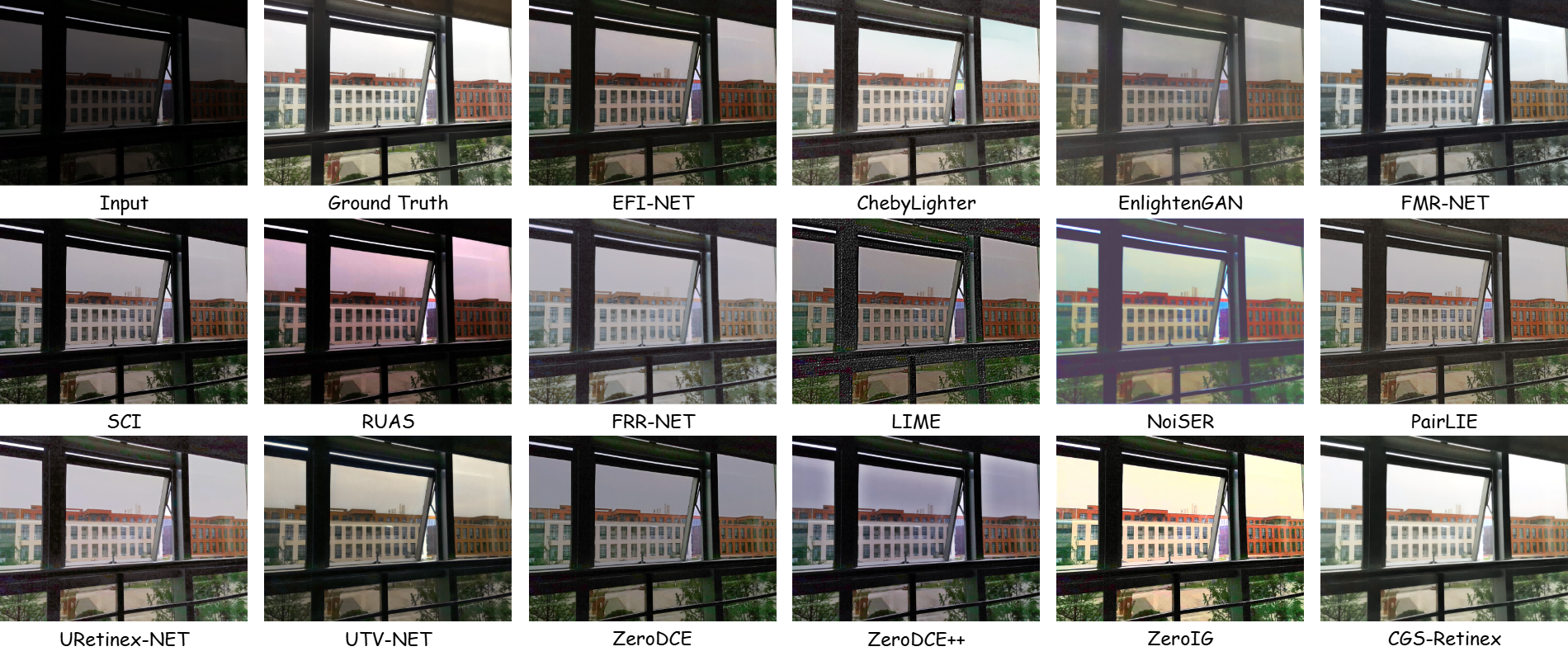}
\caption{Visual comparison of CGS-Retinex and state-of-the-art methods on the LSRW-HUAWEI dataset.}
\label{fig4}
\end{figure*}

\textbf{Implementation Details:} We implement the complete architecture of our model using the PyTorch deep learning library and execute all training and evaluation procedures on a single NVIDIA RTX 3090 GPU. During the training phase, we uniformly crop the input images into patches of size $H \times W$. To significantly enhance spatial translation invariance and scale robustness, we apply comprehensive data augmentation strategies, including random horizontal and vertical flips with a probability of 0.5. Furthermore, we incorporate a dynamic multi-scale training mechanism. With a 30\% probability, this mechanism rescales the target images via bilinear interpolation within a scaling factor interval of $[S_{min}, S_{max}]$. Consequently, this dynamic scaling explicitly forces the continuous Gaussian renderer to seamlessly adapt to multi-resolution grids.

To optimize all network parameters, we employ the AdamW optimizer with a weight decay of 1e-4 and initialize the base learning rate at 2e-4. For the learning rate decay strategy, we apply a cosine annealing scheduler with warm restarts, configuring the initial period to $T_0=50$ and the period multiplier to $T_{mult}=2$. Furthermore, to stabilize the optimization of the continuous Gaussian feature space and explicitly prevent gradient explosion, we enforce a gradient clipping threshold of 0.5. We train the entire model end-to-end for 1000 epochs utilizing a batch size of 10.

Guided by the multi-objective joint loss constraints, the comprehensive optimization objective encompasses the reconstruction loss, structural similarity loss, perceptual penalty, Total Variation (TV) illumination smoothing regularization, physics-inspired grayscale consistency constraint, and color similarity loss. Based on extensive empirical validation, we precisely configure the corresponding weight distribution defined in Equation (28) as $\lambda_1=1.0$, $\lambda_2=0.5$, $\lambda_3=0.05$, $\lambda_4=0.05$, $\lambda_5=0.05$, and $\lambda_6=0.05$. This meticulously calibrated weight architecture empowers the network to autonomously recover highly saturated colors while strictly preserving the physical consistency between the explicitly modeled illumination distribution and the implicitly represented high-frequency reflectance.

\begin{figure*}[!t]
\centering
\includegraphics[width=\linewidth]{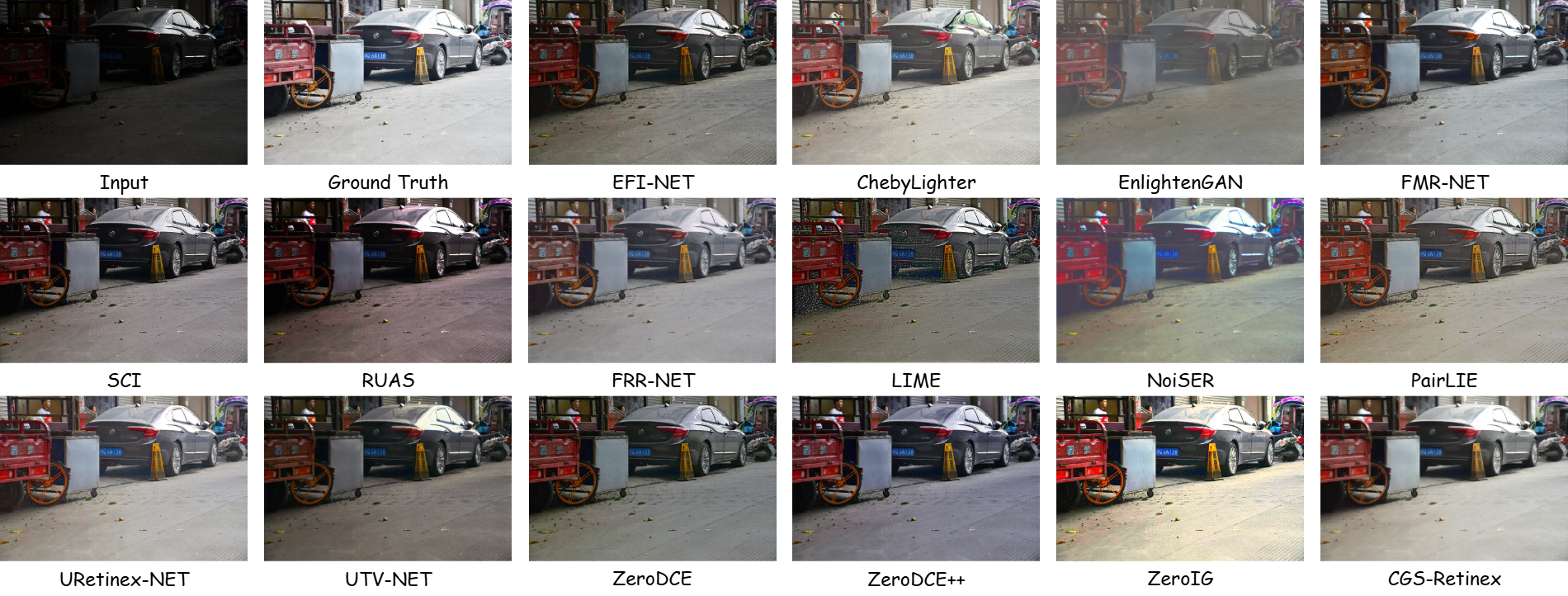}
\caption{Visual comparison of CGS-Retinex and state-of-the-art methods on the LSRW-NIKON dataset.}
\label{fig5}
\end{figure*}

\begin{figure*}[!t]
\centering
\includegraphics[width=\linewidth]{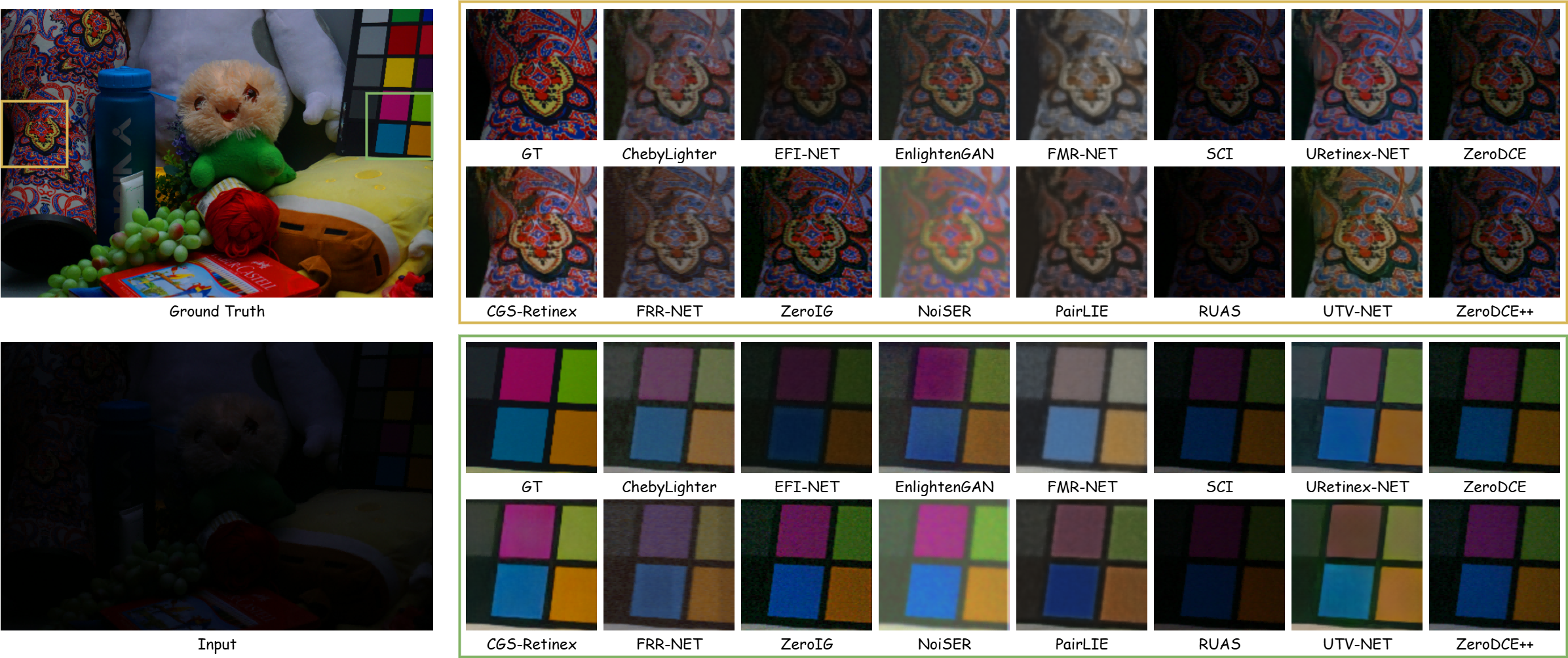}
\caption{Comparison of detailed features between CGS-Retinex and state-of-the-art methods on the LOL dataset. Magnified views of the regions highlighted by red and green boxes are provided for each method.}
\label{fig6}
\end{figure*}

\textbf{Evaluation Metrics:} To comprehensively assess the image restoration and enhancement performance, we employ seven widely recognized evaluation metrics, comprising three full-reference and four no-reference indicators. For datasets providing paired ground-truth (GT) references, we utilize three full-reference metrics: Peak Signal-to-Noise Ratio (PSNR), Structural Similarity (SSIM), and Learned Perceptual Image Patch Similarity (LPIPS). To evaluate real-world applications lacking ground-truth references, particularly for unpaired data and real-world applications, we incorporate four specialized no-reference metrics: Natural Image Quality Evaluator (NIQE), Lightness Order Error (LOE), Discrete Entropy (DE), and Enhancement Measure of Evaluation (EME).

\subsection{Comparison with State-of-the-Art Methods}
To rigorously validate the superiority of our CGS Retinex architecture, we conduct extensive comparative evaluations against recent state-of-the-art low light image enhancement methods \cite{ref4, ref5, ref7, ref8, ref9, ref10, ref16, ref17, ref18, ref19, ref22, ref39, ref40, ref41, ref42}.

\begin{table*}[!t]
\caption{Quantitative comparison on the LOL dataset. The best, second-best, and third-best results for each metric are highlighted in \colorbox{firstcolor}{orange}, \colorbox{secondcolor}{light orange}, and \colorbox{thirdcolor}{pale orange} respectively. Other values are highlighted in \colorbox{restcolor}{lightest orange}.}
\centering
\begin{tabular}{lccccccccc}
\toprule
Method & SSIM↑ & PSNR↑ & LPIPS↓ & NIQE↓ & LOE↓ & DE↑ & EME↑ & Params (KB)↓ & FLOPs (G)↓ \\
\midrule
ZeroDCE++ \cite{ref40} & \cellcolor{restcolor}0.63 & \cellcolor{restcolor}15.65 & \cellcolor{restcolor}0.31 & \cellcolor{restcolor}5.71 & \cellcolor{restcolor}27.76 & \cellcolor{restcolor}1.90 & \cellcolor{restcolor}18.56 & \cellcolor{restcolor}10.6 & \cellcolor{restcolor}0.33 \\
ZeroDCE \cite{ref39} & \cellcolor{restcolor}0.63 & \cellcolor{restcolor}15.20 & \cellcolor{restcolor}0.30 & \cellcolor{restcolor}5.66 & \cellcolor{restcolor}28.73 & \cellcolor{restcolor}1.74 & \cellcolor{restcolor}18.54 & \cellcolor{restcolor}79.4 & \cellcolor{restcolor}5.21 \\
SCI \cite{ref42} & \cellcolor{restcolor}0.59 & \cellcolor{restcolor}14.70 & \cellcolor{restcolor}0.31 & \cellcolor{restcolor}5.87 & \cellcolor{thirdcolor}9.33 & \cellcolor{restcolor}1.86 & \cellcolor{restcolor}19.27 & \cellcolor{firstcolor}0.3 & \cellcolor{firstcolor}0.0619 \\
RUAS \cite{ref16} & \cellcolor{restcolor}0.50 & \cellcolor{restcolor}12.82 & \cellcolor{restcolor}0.34 & \cellcolor{restcolor}5.70 & \cellcolor{firstcolor}0.51 & \cellcolor{restcolor}1.42 & \cellcolor{thirdcolor}20.06 & \cellcolor{secondcolor}1.4 & \cellcolor{thirdcolor}0.2813 \\
EnlightenGAN \cite{ref41} & \cellcolor{restcolor}0.72 & \cellcolor{restcolor}16.74 & \cellcolor{restcolor}0.29 & \cellcolor{restcolor}4.21 & \cellcolor{restcolor}79.08 & \cellcolor{restcolor}1.54 & \cellcolor{restcolor}3.00 & \cellcolor{restcolor}8636 & \cellcolor{restcolor}61.01 \\
FMR-NET \cite{ref7} & \cellcolor{thirdcolor}0.80 & \cellcolor{thirdcolor}19.04 & \cellcolor{restcolor}0.24 & \cellcolor{restcolor}4.57 & \cellcolor{restcolor}26.36 & \cellcolor{thirdcolor}2.18 & \cellcolor{restcolor}3.07 & \cellcolor{restcolor}196.8 & \cellcolor{restcolor}102.8 \\
LIME \cite{ref4} & \cellcolor{restcolor}0.52 & \cellcolor{restcolor}14.35 & \cellcolor{restcolor}0.39 & \cellcolor{restcolor}5.50 & \cellcolor{restcolor}103.64 & \cellcolor{restcolor}1.76 & \cellcolor{secondcolor}21.19 & \cellcolor{restcolor}N/A & \cellcolor{restcolor}N/A \\
FRR-NET \cite{ref8} & \cellcolor{restcolor}0.76 & \cellcolor{restcolor}18.18 & \cellcolor{restcolor}0.27 & \cellcolor{restcolor}5.73 & \cellcolor{restcolor}23.52 & \cellcolor{restcolor}1.84 & \cellcolor{restcolor}3.81 & \cellcolor{restcolor}12.21 & \cellcolor{secondcolor}0.216 \\
UTV-NET \cite{ref17} & \cellcolor{restcolor}0.75 & \cellcolor{restcolor}16.36 & \cellcolor{thirdcolor}0.20 & \cellcolor{restcolor}4.31 & \cellcolor{restcolor}29.06 & \cellcolor{restcolor}1.94 & \cellcolor{restcolor}4.76 & \cellcolor{restcolor}7745 & \cellcolor{restcolor}58.29 \\
ChebyLighter \cite{ref5} & \cellcolor{restcolor}0.75 & \cellcolor{restcolor}18.74 & \cellcolor{thirdcolor}0.20 & \cellcolor{restcolor}4.30 & \cellcolor{restcolor}31.72 & \cellcolor{restcolor}2.06 & \cellcolor{restcolor}3.76 & \cellcolor{restcolor}73 & \cellcolor{restcolor}17.25 \\
EFI-NET \cite{ref9} & \cellcolor{restcolor}0.67 & \cellcolor{restcolor}14.27 & \cellcolor{restcolor}0.29 & \cellcolor{restcolor}4.33 & \cellcolor{restcolor}28.74 & \cellcolor{restcolor}1.59 & \cellcolor{restcolor}5.40 & \cellcolor{restcolor}129.2 & \cellcolor{restcolor}9.38 \\
PairLIE \cite{ref10} & \cellcolor{restcolor}0.76 & \cellcolor{restcolor}18.15 & \cellcolor{restcolor}0.23 & \cellcolor{thirdcolor}4.05 & \cellcolor{restcolor}54.92 & \cellcolor{restcolor}1.81 & \cellcolor{restcolor}3.84 & \cellcolor{restcolor}34.18 & \cellcolor{restcolor}22.35 \\
NoiSER \cite{ref22} & \cellcolor{restcolor}0.70 & \cellcolor{restcolor}17.36 & \cellcolor{restcolor}0.37 & \cellcolor{restcolor}4.34 & \cellcolor{restcolor}51.61 & \cellcolor{restcolor}2.17 & \cellcolor{restcolor}2.47 & \cellcolor{thirdcolor}1.763 & \cellcolor{restcolor}8.62 \\
URetinex-NET \cite{ref18} & \cellcolor{secondcolor}0.83 & \cellcolor{secondcolor}19.93 & \cellcolor{restcolor}0.23 & \cellcolor{secondcolor}4.04 & \cellcolor{restcolor}28.76 & \cellcolor{restcolor}2.02 & \cellcolor{restcolor}3.54 & \cellcolor{restcolor}838.3 & \cellcolor{restcolor}136.01 \\
ZeroIG \cite{ref19} & \cellcolor{restcolor}0.51 & \cellcolor{restcolor}17.23 & \cellcolor{secondcolor}0.15 & \cellcolor{restcolor}6.34 & \cellcolor{restcolor}14.32 & \cellcolor{secondcolor}2.41 & \cellcolor{restcolor}19.52 & \cellcolor{restcolor}123.63 & \cellcolor{restcolor}118.73 \\
\textbf{CGS-Retinex} & \cellcolor{firstcolor}0.92 & \cellcolor{firstcolor}22.41 & \cellcolor{firstcolor}0.11 & \cellcolor{firstcolor}3.86 & \cellcolor{secondcolor}5.64 & \cellcolor{firstcolor}3.07 & \cellcolor{firstcolor}22.04 & \cellcolor{restcolor}1280 & \cellcolor{restcolor}62.73 \\
\bottomrule
\end{tabular}
\end{table*}

\begin{table*}[!t]
\caption{Quantitative comparison on the LSRW-HUAWEI dataset. Colors as defined in Table I.}
\centering
\begin{tabular}{lccccccccc}
\toprule
Method & SSIM↑ & PSNR↑ & LPIPS↓ & NIQE↓ & LOE↓ & DE↑ & EME↑ & Params (KB)↓ & FLOPs (G)↓ \\
\midrule
ZeroDCE++ \cite{ref40} & \cellcolor{restcolor}0.51 & \cellcolor{restcolor}15.14 & \cellcolor{restcolor}0.38 & \cellcolor{restcolor}3.18 & \cellcolor{restcolor}19.29 & \cellcolor{restcolor}2.02 & \cellcolor{restcolor}18.54 & \cellcolor{restcolor}10.6 & \cellcolor{restcolor}0.33 \\
ZeroDCE \cite{ref39} & \cellcolor{restcolor}0.51 & \cellcolor{restcolor}14.79 & \cellcolor{restcolor}0.36 & \cellcolor{restcolor}3.16 & \cellcolor{restcolor}20.51 & \cellcolor{restcolor}1.94 & \cellcolor{restcolor}18.42 & \cellcolor{restcolor}79.4 & \cellcolor{restcolor}5.21 \\
SCI \cite{ref42} & \cellcolor{restcolor}0.45 & \cellcolor{restcolor}13.69 & \cellcolor{restcolor}0.39 & \cellcolor{restcolor}3.42 & \cellcolor{thirdcolor}4.86 & \cellcolor{restcolor}1.88 & \cellcolor{restcolor}19.98 & \cellcolor{firstcolor}0.3 & \cellcolor{firstcolor}0.0619 \\
RUAS \cite{ref16} & \cellcolor{restcolor}0.34 & \cellcolor{restcolor}11.46 & \cellcolor{restcolor}0.45 & \cellcolor{restcolor}3.70 & \cellcolor{firstcolor}0.85 & \cellcolor{restcolor}1.38 & \cellcolor{thirdcolor}20.96 & \cellcolor{secondcolor}1.4 & \cellcolor{thirdcolor}0.2813 \\
EnlightenGAN \cite{ref41} & \cellcolor{restcolor}0.58 & \cellcolor{restcolor}17.10 & \cellcolor{restcolor}0.34 & \cellcolor{restcolor}3.24 & \cellcolor{restcolor}58.86 & \cellcolor{restcolor}1.54 & \cellcolor{restcolor}2.33 & \cellcolor{restcolor}8636 & \cellcolor{restcolor}61.01 \\
FMR-NET \cite{ref7} & \cellcolor{secondcolor}0.64 & \cellcolor{thirdcolor}19.66 & \cellcolor{restcolor}0.33 & \cellcolor{restcolor}3.86 & \cellcolor{restcolor}21.30 & \cellcolor{thirdcolor}2.34 & \cellcolor{restcolor}3.23 & \cellcolor{restcolor}196.8 & \cellcolor{restcolor}102.8 \\
LIME \cite{ref4} & \cellcolor{restcolor}0.37 & \cellcolor{restcolor}13.71 & \cellcolor{restcolor}0.49 & \cellcolor{restcolor}3.96 & \cellcolor{restcolor}109.63 & \cellcolor{restcolor}1.95 & \cellcolor{secondcolor}22.01 & \cellcolor{restcolor}N/A & \cellcolor{restcolor}N/A \\
FRR-NET \cite{ref8} & \cellcolor{restcolor}0.58 & \cellcolor{restcolor}18.76 & \cellcolor{restcolor}0.36 & \cellcolor{restcolor}3.60 & \cellcolor{restcolor}15.17 & \cellcolor{restcolor}2.13 & \cellcolor{restcolor}4.45 & \cellcolor{restcolor}12.21 & \cellcolor{secondcolor}0.216 \\
UTV-NET \cite{ref17} & \cellcolor{thirdcolor}0.62 & \cellcolor{restcolor}17.86 & \cellcolor{firstcolor}0.25 & \cellcolor{restcolor}3.12 & \cellcolor{restcolor}19.42 & \cellcolor{restcolor}2.29 & \cellcolor{restcolor}5.00 & \cellcolor{restcolor}7745 & \cellcolor{restcolor}58.29 \\
ChebyLighter \cite{ref5} & \cellcolor{restcolor}0.61 & \cellcolor{restcolor}18.74 & \cellcolor{restcolor}0.30 & \cellcolor{firstcolor}2.77 & \cellcolor{restcolor}20.97 & \cellcolor{restcolor}2.25 & \cellcolor{restcolor}4.16 & \cellcolor{restcolor}73 & \cellcolor{restcolor}17.25 \\
EFI-NET \cite{ref9} & \cellcolor{restcolor}0.51 & \cellcolor{restcolor}13.52 & \cellcolor{restcolor}0.37 & \cellcolor{restcolor}3.47 & \cellcolor{restcolor}22.54 & \cellcolor{restcolor}1.69 & \cellcolor{restcolor}5.92 & \cellcolor{restcolor}129.2 & \cellcolor{restcolor}9.38 \\
PairLIE \cite{ref10} & \cellcolor{restcolor}0.61 & \cellcolor{restcolor}18.09 & \cellcolor{restcolor}0.32 & \cellcolor{restcolor}3.53 & \cellcolor{restcolor}47.1 & \cellcolor{restcolor}1.97 & \cellcolor{restcolor}3.76 & \cellcolor{restcolor}34.18 & \cellcolor{restcolor}22.35 \\
NoiSER \cite{ref22} & \cellcolor{restcolor}0.58 & \cellcolor{restcolor}16.31 & \cellcolor{restcolor}0.49 & \cellcolor{restcolor}4.18 & \cellcolor{restcolor}37.05 & \cellcolor{restcolor}2.03 & \cellcolor{restcolor}1.90 & \cellcolor{thirdcolor}1.763 & \cellcolor{restcolor}8.62 \\
URetinex-NET \cite{ref18} & \cellcolor{thirdcolor}0.62 & \cellcolor{secondcolor}19.91 & \cellcolor{thirdcolor}0.28 & \cellcolor{thirdcolor}3.07 & \cellcolor{restcolor}20.92 & \cellcolor{restcolor}2.30 & \cellcolor{restcolor}4.66 & \cellcolor{restcolor}838.3 & \cellcolor{restcolor}136.01 \\
ZeroIG \cite{ref19} & \cellcolor{restcolor}0.48 & \cellcolor{restcolor}17.29 & \cellcolor{restcolor}0.40 & \cellcolor{restcolor}3.51 & \cellcolor{restcolor}9.42 & \cellcolor{secondcolor}2.45 & \cellcolor{restcolor}19.17 & \cellcolor{restcolor}123.63 & \cellcolor{restcolor}118.73 \\
\textbf{CGS-Retinex} & \cellcolor{firstcolor}0.87 & \cellcolor{firstcolor}20.19 & \cellcolor{secondcolor}0.27 & \cellcolor{secondcolor}3.06 & \cellcolor{secondcolor}4.30 & \cellcolor{firstcolor}2.85 & \cellcolor{firstcolor}22.71 & \cellcolor{restcolor}1280 & \cellcolor{restcolor}62.73 \\
\bottomrule
\end{tabular}
\end{table*}

% --- 表格 3：跨双栏 ---
\begin{table*}[!t]
\caption{Quantitative comparison on the LSRW-NIKON dataset. Colors as defined in Table I.}
\centering
\begin{tabular}{lccccccccc}
\toprule
Method & SSIM↑ & PSNR↑ & LPIPS↓ & NIQE↓ & LOE↓ & DE↑ & EME↑ & Params (KB)↓ & FLOPs (G)↓ \\
\midrule
ZeroDCE++ \cite{ref40} & \cellcolor{restcolor}0.47 & \cellcolor{restcolor}16.16 & \cellcolor{restcolor}0.39 & \cellcolor{restcolor}3.78 & \cellcolor{restcolor}38.61 & \cellcolor{restcolor}1.59 & \cellcolor{restcolor}13.91 & \cellcolor{restcolor}10.6 & \cellcolor{restcolor}0.33 \\
ZeroDCE \cite{ref39} & \cellcolor{restcolor}0.45 & \cellcolor{restcolor}15.55 & \cellcolor{restcolor}0.38 & \cellcolor{restcolor}3.78 & \cellcolor{restcolor}32.02 & \cellcolor{restcolor}1.43 & \cellcolor{restcolor}13.88 & \cellcolor{restcolor}79.4 & \cellcolor{restcolor}5.21 \\
SCI \cite{ref42} & \cellcolor{restcolor}0.41 & \cellcolor{restcolor}15.82 & \cellcolor{restcolor}0.39 & \cellcolor{restcolor}3.74 & \cellcolor{thirdcolor}12.81 & \cellcolor{restcolor}1.61 & \cellcolor{restcolor}15.2 & \cellcolor{firstcolor}0.3 & \cellcolor{firstcolor}0.0619 \\
RUAS \cite{ref16} & \cellcolor{restcolor}0.38 & \cellcolor{restcolor}13.02 & \cellcolor{restcolor}0.42 & \cellcolor{restcolor}3.97 & \cellcolor{firstcolor}1.62 & \cellcolor{restcolor}1.21 & \cellcolor{thirdcolor}16.01 & \cellcolor{secondcolor}1.4 & \cellcolor{thirdcolor}0.2813 \\
EnlightenGAN \cite{ref41} & \cellcolor{restcolor}0.46 & \cellcolor{restcolor}15.30 & \cellcolor{restcolor}0.40 & \cellcolor{restcolor}3.54 & \cellcolor{restcolor}79.60 & \cellcolor{restcolor}0.94 & \cellcolor{restcolor}2.55 & \cellcolor{restcolor}8636 & \cellcolor{restcolor}61.01 \\
FMR-NET \cite{ref7} & \cellcolor{secondcolor}0.75 & \cellcolor{secondcolor}18.47 & \cellcolor{thirdcolor}0.35 & \cellcolor{restcolor}3.91 & \cellcolor{restcolor}20.71 & \cellcolor{restcolor}1.70 & \cellcolor{restcolor}3.31 & \cellcolor{restcolor}196.8 & \cellcolor{restcolor}102.8 \\
LIME \cite{ref4} & \cellcolor{restcolor}0.30 & \cellcolor{restcolor}14.51 & \cellcolor{restcolor}0.51 & \cellcolor{restcolor}4.29 & \cellcolor{restcolor}90.6 & \cellcolor{restcolor}1.55 & \cellcolor{firstcolor}17.04 & \cellcolor{restcolor}N/A & \cellcolor{restcolor}N/A \\
FRR-NET \cite{ref8} & \cellcolor{restcolor}0.52 & \cellcolor{restcolor}17.34 & \cellcolor{restcolor}0.36 & \cellcolor{restcolor}4.45 & \cellcolor{restcolor}22.46 & \cellcolor{restcolor}1.51 & \cellcolor{restcolor}3.51 & \cellcolor{restcolor}12.21 & \cellcolor{secondcolor}0.216 \\
UTV-NET \cite{ref17} & \cellcolor{restcolor}0.53 & \cellcolor{restcolor}15.94 & \cellcolor{firstcolor}0.32 & \cellcolor{restcolor}4.00 & \cellcolor{restcolor}21.11 & \cellcolor{restcolor}1.55 & \cellcolor{restcolor}5.12 & \cellcolor{restcolor}7745 & \cellcolor{restcolor}58.29 \\
ChebyLighter \cite{ref5} & \cellcolor{restcolor}0.46 & \cellcolor{restcolor}15.68 & \cellcolor{restcolor}0.37 & \cellcolor{restcolor}3.12 & \cellcolor{restcolor}60.95 & \cellcolor{thirdcolor}1.75 & \cellcolor{restcolor}4.48 & \cellcolor{restcolor}73 & \cellcolor{restcolor}17.25 \\
EFI-NET \cite{ref9} & \cellcolor{restcolor}0.47 & \cellcolor{restcolor}15.25 & \cellcolor{restcolor}0.36 & \cellcolor{restcolor}3.69 & \cellcolor{restcolor}25.16 & \cellcolor{restcolor}1.43 & \cellcolor{restcolor}5.63 & \cellcolor{restcolor}129.2 & \cellcolor{restcolor}9.38 \\
PairLIE \cite{ref10} & \cellcolor{restcolor}0.69 & \cellcolor{thirdcolor}18.33 & \cellcolor{restcolor}0.36 & \cellcolor{thirdcolor}3.07 & \cellcolor{restcolor}44.56 & \cellcolor{restcolor}1.49 & \cellcolor{restcolor}3.73 & \cellcolor{restcolor}34.18 & \cellcolor{restcolor}22.35 \\
NoiSER \cite{ref22} & \cellcolor{restcolor}0.50 & \cellcolor{restcolor}15.89 & \cellcolor{restcolor}0.53 & \cellcolor{restcolor}3.90 & \cellcolor{restcolor}35.06 & \cellcolor{restcolor}1.26 & \cellcolor{restcolor}1.91 & \cellcolor{thirdcolor}1.763 & \cellcolor{restcolor}8.62 \\
URetinex-NET \cite{ref18} & \cellcolor{thirdcolor}0.70 & \cellcolor{restcolor}17.95 & \cellcolor{secondcolor}0.34 & \cellcolor{firstcolor}2.98 & \cellcolor{restcolor}21.99 & \cellcolor{restcolor}1.67 & \cellcolor{restcolor}3.95 & \cellcolor{restcolor}838.3 & \cellcolor{restcolor}136.01 \\
ZeroIG \cite{ref19} & \cellcolor{restcolor}0.35 & \cellcolor{restcolor}15.74 & \cellcolor{restcolor}0.46 & \cellcolor{restcolor}4.11 & \cellcolor{restcolor}19.84 & \cellcolor{secondcolor}1.86 & \cellcolor{restcolor}14.65 & \cellcolor{restcolor}123.63 & \cellcolor{restcolor}118.73 \\
\textbf{CGS-Retinex} & \cellcolor{firstcolor}0.82 & \cellcolor{firstcolor}19.97 & \cellcolor{secondcolor}0.34 & \cellcolor{secondcolor}3.04 & \cellcolor{secondcolor}11.08 & \cellcolor{firstcolor}1.91 & \cellcolor{secondcolor}16.52 & \cellcolor{restcolor}1280 & \cellcolor{restcolor}62.73 \\
\bottomrule
\end{tabular}
\end{table*}

Visual comparison results on the LOL dataset are presented in Fig. \ref{fig3}. In terms of global illumination restoration relative to the ground truth (GT), the highest fidelity is achieved by ChebyLighter, FMR-NET, ZeroIG, and CGS-Retinex. However, the enhanced results of NoiSER are degraded by excessive illumination. Conversely, severe illumination deficiency is observed in several algorithms such as RUAS, EFI-NET, LIME, and ZeroDCE. In terms of color restoration, CGS-Retinex achieves competitive performance without exhibiting any color deviation. Conversely, algorithms including FMR-NET, UTV-NET, and EnlightenGAN suffer from global color shifts to varying degrees. Meanwhile, NoiSER and ZeroIG are affected by certain levels of over-saturation. Regarding the overall clarity, the enhanced results of EnlightenGAN, NoiSER, and PairLIE are degraded by visible haziness and blurred details. In contrast, CGS-Retinex demonstrates superior visual quality, which is characterized by realistic colors, appropriate illumination, and sharp structural representations.

Visual comparison results on the LSRW-HUAWEI dataset are illustrated in Fig. \ref{fig4}. In terms of illumination restoration and color fidelity, several algorithms such as RUAS, LIME, ZeroDCE, and UTV-NET suffer from suboptimal brightness. Although NoiZER, ZeroIG, ChebyLighter, and URetinex-NET effectively enhance the illumination, these methods are affected by noticeable color deviations. Conversely, CGS-Retinex maintains appropriate brightness while achieving color restoration highly consistent with the ground truth (GT). Regarding the overall clarity, the enhanced results of EnlightenGAN, ZeroDCE, RUAS, and FRR-NET are degraded by varying degrees of haziness and blurring artifacts. In contrast, CGS-Retinex demonstrates superior performance characterized by sharp structural representations and visually pleasing outcomes.

Visual comparison results on the LSRW-NIKON dataset are illustrated in Fig. \ref{fig5}. In terms of global illumination enhancement, the most competitive performance is achieved by ChebyLighter, URetinex-NET, ZeroIG, and CGS-Retinex. However, the enhanced results of ChebyLighter are affected by visible haziness and blurring artifacts. Meanwhile, ZeroIG is degraded by excessive contrast and noticeable color deviations. In contrast, CGS-Retinex demonstrates superior performance, which is characterized by moderate contrast, appropriate illumination level, and high color fidelity consistent with the ground truth (GT).

To further evaluate the superiority of CGS-Retinex in detail restoration, visual comparisons on complex scenes from the LOL dataset are presented in Fig. \ref{fig6}. The upper part of the figure focuses on the restoration of intricate patterns, while the lower part is utilized to verify image sharpness and denoising efficacy. In the upper examples, the results produced by CGS-Retinex achieve the highest consistency with the ground truth (GT) regarding illumination, color fidelity, and texture details. In terms of the lower comparisons, the proposed method demonstrates the most effective noise suppression among all evaluated algorithms, despite a minor discrepancy in saturation compared to the GT.

In summary, CGS-Retinex achieves state-of-the-art (SOTA) performance in terms of subjective visual metrics, including illumination, saturation, detail restoration, and clarity. Benefiting from the precise decoupling of illumination and texture, CGS-Retinex also demonstrates superior efficacy in noise suppression and overexposure mitigation.

Quantitative results across three benchmarks are summarized in Tables I, II, and III. Although CGS-Retinex is not optimized for parameter efficiency or FLOPs, the highest scores in PSNR and SSIM are achieved by our framework on all evaluated datasets. Regarding the remaining metrics, CGS-Retinex consistently ranks within the top two positions. Since the proposed method is established as a pioneering modeling paradigm rather than a lightweight architecture, it effectively strikes a balance between image quality and methodological innovation. Consequently, a clear competitive advantage is demonstrated by CGS-Retinex compared with existing state-of-the-art (SOTA) methods.

\subsection{Ablation Study}
To validate the effectiveness of the key components and loss functions in the proposed CGS-Retinex, extensive ablation studies are conducted on the LOL dataset. To ensure a fair comparison, all variants are trained using identical strategies and hyperparameter configurations as the full model.
% --- 表格 4：单栏 ---
\begin{table}[!t]
\caption{Quantitative validation of the illumination-reflectance formation paradigm.}
\centering
\begin{tabular}{lccc}
\toprule
Formulation & Residual Compensation & SSIM$\uparrow$ & PSNR$\uparrow$ \\
\midrule
$Out=\mathcal{F}(I_{in})$ & $\times$ & \cellcolor{thirdcolor}0.74 & \cellcolor{thirdcolor}18.17 \\
$Out=I_{in}/\mathcal{L}$ & $\times$ & \cellcolor{secondcolor}0.82 & \cellcolor{secondcolor}18.47 \\
$Out=I_{in}/\mathcal{L}+\mathcal{R}_{res}$ & \checkmark & \cellcolor{firstcolor}0.92 & \cellcolor{firstcolor}22.41 \\
\bottomrule
\end{tabular}
\end{table}

\begin{table}[!t]
\caption{Quantitative validation of the continuous splatting renderer and adaptive temperature.}
\centering
\begin{tabular}{llcc}
\toprule
Decoder Type & Setting & SSIM$\uparrow$ & PSNR$\uparrow$ \\
\midrule
Bilinear+Conv & N/A & \cellcolor{restcolor}0.76 & \cellcolor{restcolor}18.74 \\
$\mathcal{R}_{cgs}$ & $T=1.0$ & \cellcolor{secondcolor}0.81 & \cellcolor{secondcolor}19.82 \\
$\mathcal{R}_{cgs}$ & $T=10.0$ & \cellcolor{thirdcolor}0.79 & \cellcolor{thirdcolor}19.70 \\
$\mathcal{R}_{cgs}$ & Learnable & \cellcolor{firstcolor}0.92 & \cellcolor{firstcolor}22.41 \\
\bottomrule
\end{tabular}
\end{table}
\textbf{Effectiveness of the Illumination-Reflectance Formation Paradigm:} In conventional Retinex models, the reflectance map is typically estimated directly or the physical constraint $S = I \cdot \mathcal{R}$ is assumed. A novel formulation is proposed to enable the network to perform fundamental illumination division initially. Subsequently, high-frequency textures and colors are compensated via residual representations. As defined in Eq. (6), this process is mathematically simplified as:
\begin{equation}
Out = \frac{I_{in}}{\mathcal{L}} + \mathcal{R}_{res}
\end{equation}

Three primary variants are compared in this experiment. For the first variant, an encoder-decoder architecture is utilized to directly map low-light inputs to enhanced images without illumination-residual decoupling. In the second variant, only the illumination map $\mathcal{L}$ is predicted and the output is strictly defined as $Out = I_{in} / \mathcal{L}$. The third variant corresponds to the proposed CGS-Retinex architecture.

Quantitative results for these variants are summarized in Table 4. The lowest performance is yielded by the first variant due to the inability of direct mapping to address severe color attenuation under extreme low-light conditions. Although a physical model is incorporated in the second variant, severe noise amplification is observed in extremely dark regions. By decoupling the base illumination from high-frequency and color residuals, a PSNR improvement of more than 2 dB is achieved by the proposed architecture. Consequently, the superiority of this paradigm in detail preservation and noise suppression is fully validated.

\textbf{Continuous Splatting Renderer and Adaptive Temperature:} As one of the core contributions, a continuous Gaussian splatting renderer $\mathcal{R}_{cgs}$ based on neighborhood blending is proposed. Furthermore, a learnable temperature coefficient is introduced to regulate the sharpness of alpha compositing. To validate the effectiveness of these designs, four specific variants are evaluated. For the first variant, $\mathcal{R}_{cgs}$ is removed and replaced by bilinear upsampling operations and convolutional layers to decode the illumination and features. In the second variant, $\mathcal{R}_{cgs}$ is retained while the learnable temperature is fixed to a standard Softmax distribution ($T=1.0$). Consequently, excessively smooth blending results are yielded by this configuration. Regarding the third variant, the temperature is rigidly fixed at $T=10.0$ to perform hard alpha compositing. Severe gradient truncation and structural artifacts are frequently induced under this setting. The fourth variant corresponds to the complete CGS-Retinex architecture. As defined in Eq. (15), a Softplus function is utilized by this full model to adaptively learn the distribution parameters.

Quantitative results are summarized in Table 5. Significant performance gains across all metrics are achieved by replacing conventional convolutional upsampling operations with $\mathcal{R}_{cgs}$. Furthermore, the diverse sharpness requirements of textures under varying illumination conditions cannot be accommodated by the fixed temperature settings in the second and third variants. By utilizing the proposed learnable temperature mechanism, an optimal balance between noise suppression and edge preservation is adaptively achieved, yielding superior subjective visual quality and objective evaluations.

\textbf{Internal Structure of Coordinate INR $\Phi_{INR}$:} To generate high-quality residuals $\mathcal{R}_{res}$, a coordinate-driven implicit neural representation network denoted as $\Phi_{INR}$ is employed. High-frequency positional encodings $\gamma(p)$ and skip connections $\mathcal{F}_{skip}$ from the encoder are primarily integrated within this module. To validate the effectiveness of this structure, four specific architectural variants are evaluated. For the first variant, $\Phi_{INR}$ is removed and replaced by a conventional feature-based convolutional network to generate the residuals. In the second variant, the high-frequency Fourier basis $\gamma(p)$ is omitted and raw $(x, y)$ coordinates are directly utilized as inputs. Regarding the third variant, the concatenation of skip connections $\mathcal{F}_{skip}(p)$ from the first encoder layer is removed. The fourth variant corresponds to the complete $\Phi_{INR}$ configuration.

Table 6 demonstrates that the complete $\Phi_{INR}$ configuration yields substantial improvements across all performance metrics. The removal of skip connections $\mathcal{F}_{skip}(p)$ results in the loss of spatially aligned fine-grained contextual features and induces blurred edges. Furthermore, the omission of positional encoding $\gamma(p)$ hinders the restoration of high-frequency details and leads to a significant decline in performance.

% --- 表格 6：单栏 ---
\begin{table}[!t]
\caption{Quantitative validation of the internal structure of the coordinate INR.}
\centering
\begin{tabular}{lcccc}
\toprule
Representation & $\gamma(p)$ & $\mathcal{F}_{skip}(p)$ & SSIM$\uparrow$ & PSNR$\uparrow$ \\
\midrule
CNN-based & N/A & \checkmark & \cellcolor{restcolor}0.82 & \cellcolor{restcolor}19.84 \\
$\Phi_{INR}$ & $\times$ & \checkmark & \cellcolor{thirdcolor}0.85 & \cellcolor{thirdcolor}20.93 \\
$\Phi_{INR}$ & \checkmark & $\times$ & \cellcolor{secondcolor}0.88 & \cellcolor{secondcolor}21.26 \\
$\Phi_{INR}$ & \checkmark & \checkmark & \cellcolor{firstcolor}0.92 & \cellcolor{firstcolor}22.41 \\
\bottomrule
\end{tabular}
\end{table}

\textbf{Disentanglement of Physical Consistency and Color Loss:} Within the loss function design, a physical consistency loss $\mathcal{L}_{phy}$ is proposed to constrain grayscale illumination. Simultaneously, higher degrees of freedom for color restoration are granted to the network and hue consistency is maintained via a cosine similarity color loss $\mathcal{L}_{color}$. Ablation results for these two loss components are summarized in Table 7. Consequently, the effectiveness of reducing physical constraints to the grayscale domain in conjunction with explicit color vector supervision is successfully validated.

% --- 表格 7：单栏 ---
\begin{table}[!t]
\caption{Quantitative validation of the disentanglement of physical consistency and color loss.}
\centering
\begin{tabular}{llcc}
\toprule
$\mathcal{L}_{phy}$ & $\mathcal{L}_{color}$ & SSIM$\uparrow$ & PSNR$\uparrow$ \\
\midrule
$\times$ & \checkmark & \cellcolor{restcolor}0.76 & \cellcolor{restcolor}19.77 \\
RGB Channels & \checkmark & \cellcolor{thirdcolor}0.78 & \cellcolor{thirdcolor}20.94 \\
Grayscale & $\times$ & \cellcolor{secondcolor}0.89 & \cellcolor{secondcolor}21.78 \\
Grayscale & \checkmark & \cellcolor{firstcolor}0.92 & \cellcolor{firstcolor}22.41 \\
\bottomrule
\end{tabular}
\end{table}

\subsection{Limitations and Future Work}
Although superior enhancement performance and visual fidelity are achieved by CGS-Retinex in low-light scenarios, certain limitations persist. First, higher memory consumption and computational overhead are introduced during high-resolution image processing compared with conventional CNNs. This limitation originates from the integration of dense neighborhood aggregation within the continuous Gaussian renderer and pixel-wise coordinate-based implicit neural representations. Second, a risk of overfitting is induced by the absence of effective feature priors in extremely dark regions or areas corrupted by severe shot noise. Consequently, noise-induced artifacts or minor texture hallucinations are occasionally generated by the INR module.

Future efforts will be directed towards two primary aspects. First, lightweight sparse Gaussian splatting mechanisms and customized CUDA kernels will be explored to significantly reduce computational complexity, thereby facilitating real-time inference on mobile platforms. Second, the proposed explicit-implicit physical representation paradigm will be extended to video low-light enhancement tasks. By introducing temporal Gaussian consistency constraints, enhancement stability and spatio-temporal coherence in dynamic scenes are expected to be further improved.

% --- 第五章：Conclusion ---
\section{Conclusion}
In this paper, a pioneering joint explicit-implicit modeling framework denoted as CGS-Retinex is proposed. By deeply integrating continuous Gaussian splatting with Retinex theory, a novel continuous physical representation paradigm is established for low-light image enhancement. Through the construction of a continuous Gaussian renderer with adaptive temperature regulation, spatial artifacts inherent in traditional discrete grids are fundamentally eliminated. Consequently, smooth and physically accurate global illumination estimation is successfully achieved. Furthermore, with the guidance of shallow high-frequency features, the implicit neural representation exhibits superior efficacy in compensating for degraded textures and colors within the residual space. Through the integration of physically-inspired illumination consistency constraints, explicit illumination and implicit reflectance are effectively decoupled. Extensive experiments demonstrate that an optimal balance among noise suppression in dark regions, overexposure mitigation, and structural detail preservation is achieved by CGS-Retinex. This framework establishes a significant theoretical perspective and application paradigm for high-order reconstruction tasks in low-level vision.

% --- 附加信息部分 ---
\section*{Acknowledgement}
During the preparation of this manuscript, the author used Google Gemini (Version 3.1 Pro) for the purposes of improving sentence flow and academic tone. The author has reviewed and edited the output and takes full responsibility for the content of this publication.

\section*{CRediT authorship contribution statement}
\textbf{Yuhan Chen:} Conceptualization, Methodology, Visualization, Writing - original draft, Writing - review \& editing, Resources, Formal analysis, Software, Validation. \textbf{Yicui Shi:} Data curation, Investigation, Software, Writing - review \& editing. \textbf{Guofa Li:} Methodology, Funding acquisition, Supervision, Resources, Writing - review \& editing. \textbf{Guangrui Bai:} Methodology, Writing - review \& editing. \textbf{Wenxuan Yu:} Software, Writing - review \& editing. \textbf{Ying Fang:} Visualization, Writing - review \& editing. \textbf{Wenbo Chu:} Resources, Writing - review \& editing. \textbf{Keqiang Li:} Resources, Writing - review \& editing.

\begin{IEEEbiography}[{\includegraphics[width=1in,height=1.25in,clip,keepaspectratio]{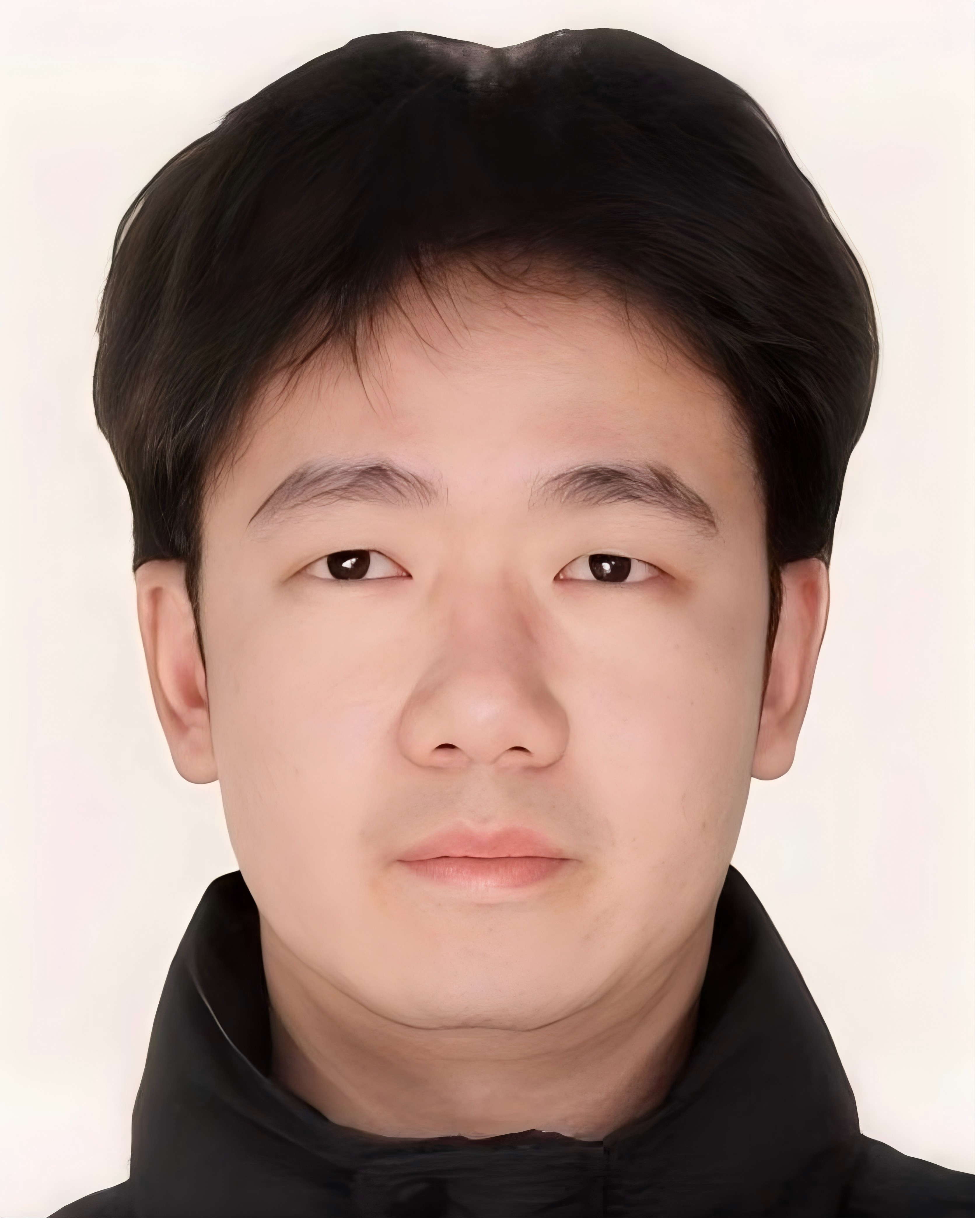}}]{Yuhan Chen}
Yuhan Chen received his master's degree in 2024 from the College of Mechanical Engineering at Chongqing University of Technology. He is currently pursuing the Ph.D. degree in College of Mechanical and Vehicle Engineering at Chongqing University, China. His research interests include deep learning, Low-level Vision and Gaussian Splatting.
\end{IEEEbiography}
\vspace{-50pt}

\begin{IEEEbiography}[{\includegraphics[width=1in,height=1.25in,clip,keepaspectratio]{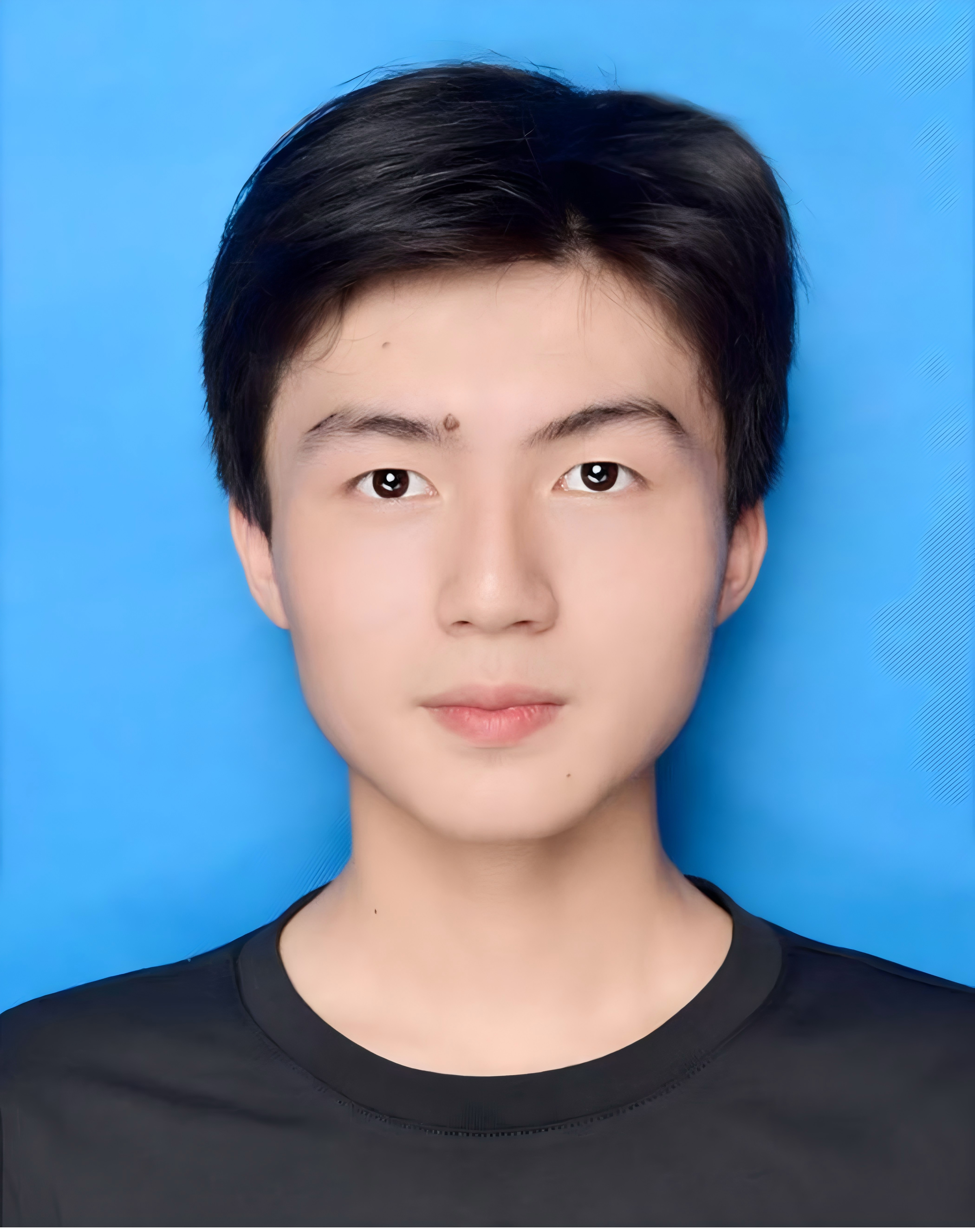}}]{Yicui Shi}
Yicui Shi received the B.E. degree majoring in Automotive Engineering at Chongqing University in 2025. He is currently pursuing the M.E. degree in Automotive Engineering at Chongqing University, Chongqing, China. His research interests include computer vision, Gaussian Splatting and deep learning.
\end{IEEEbiography}
\vspace{-40pt}

\begin{IEEEbiography}[{\includegraphics[width=1in,height=1.25in,clip,keepaspectratio]{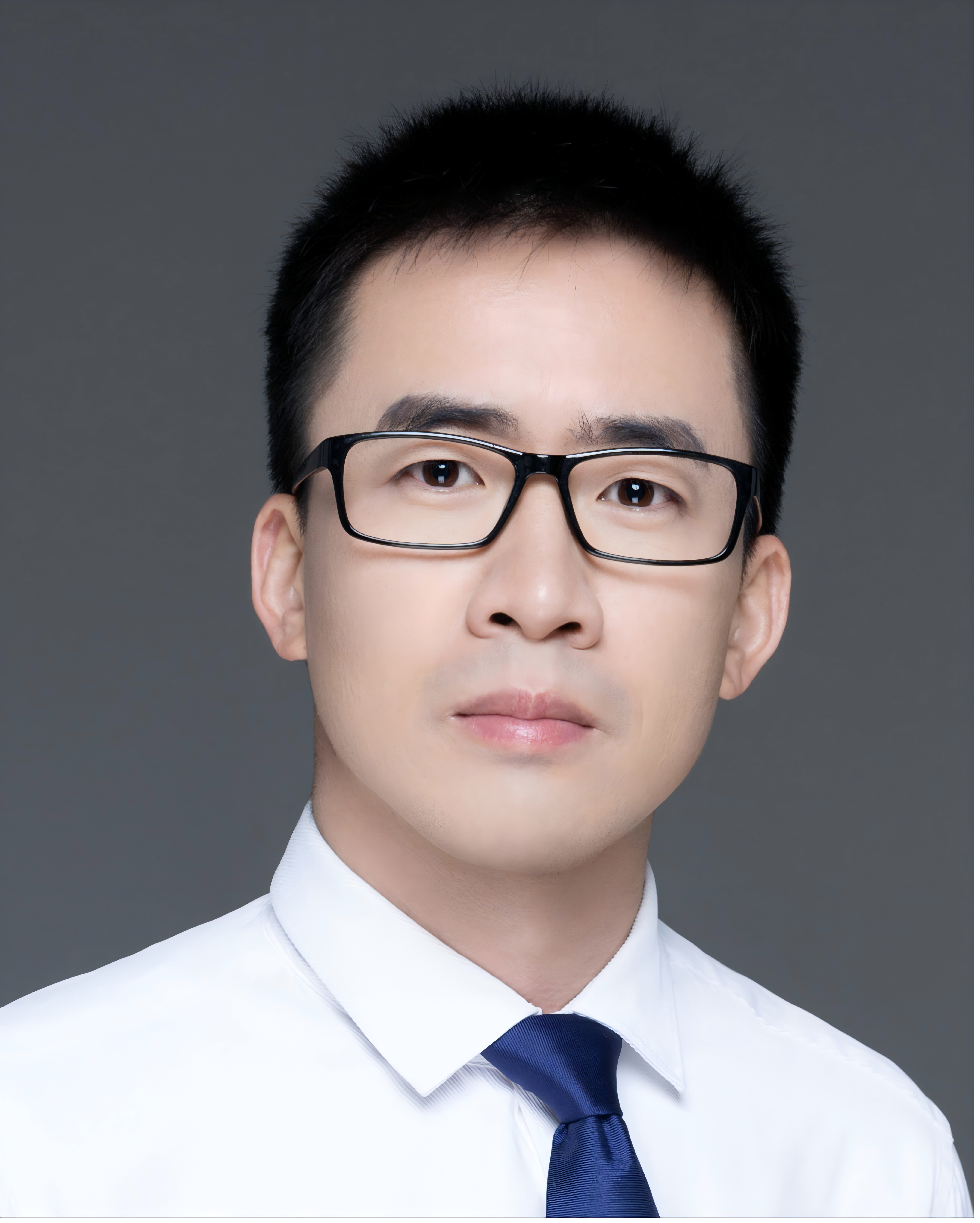}}]{Guofa Li}
Guofa Li received the Ph.D. degree in Mechanical Engineering from Tsinghua University, China, in 2016. He is currently a Professor with Chongqing University, China. His research interests include environment perception, driver behavior analysis, and smart decision-making based on artificial intelligence technologies in autonomous vehicles and intelligent transportation systems. He serves as the Associate Editor for IEEE Transactions on Intelligent Transportation Systems, IEEE Transactions on Affective Computing, and IEEE Sensors Journal.
\end{IEEEbiography}
\vspace{-40pt}

\begin{IEEEbiography}[{\includegraphics[width=1in,height=1.25in,clip,keepaspectratio]{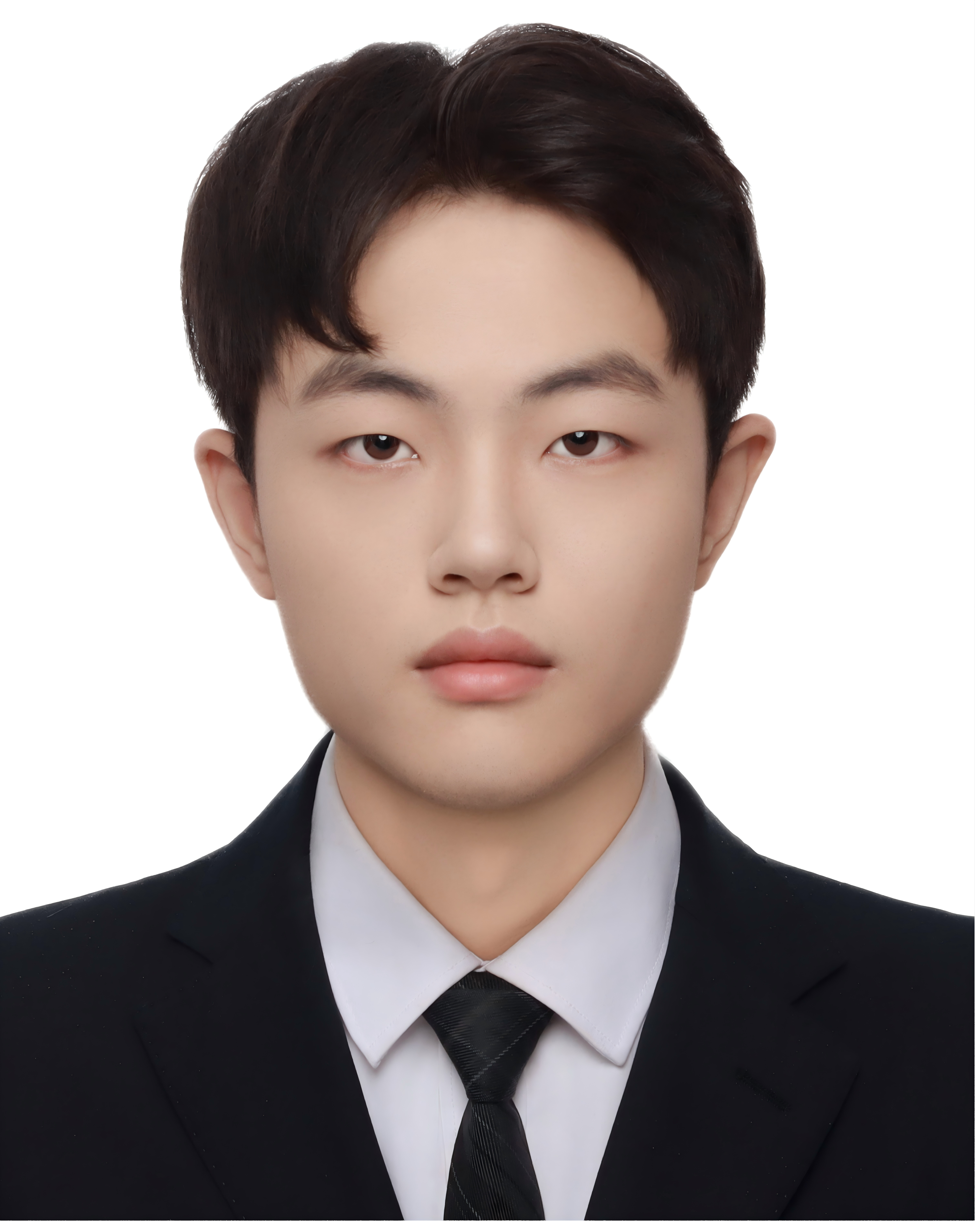}}]{Wenxuan Yu}
Wenxuan Yu received the B.E. degree majoring in Mechanical Design, Manufacturing, and Automation at Chongqing University in 2025. He is currently pursuing the M.E. degree in Mechanical Engineering at Chongqing University, Chongqing, China. His research interests include computer vision, Gaussian Splatting and deep learning.
\end{IEEEbiography}
\vspace{-40pt}

\begin{IEEEbiography}[{\includegraphics[width=1in,height=1.25in,clip,keepaspectratio]{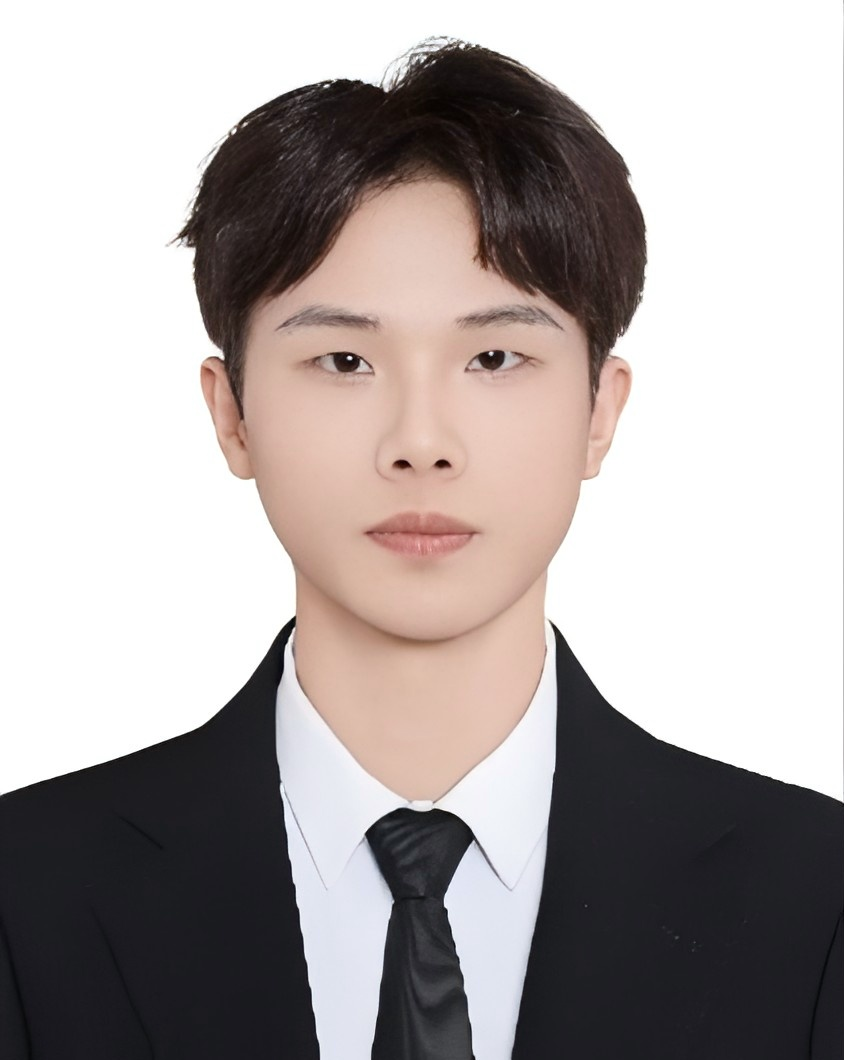}}]{Ying Fang}
Ying Fang received the B.E. degree majoring in Vehicle Engineering at Chongqing University of Technology. He is currently pursuing the M.E. degree in Mechanical Engineering at Chongqing University, Chongqing, China. His research interests include computer vision, Gaussian Splatting and deep learning.
\end{IEEEbiography}
\vspace{-40pt}

\begin{IEEEbiography}[{\includegraphics[width=1in,height=1.25in,clip,keepaspectratio]{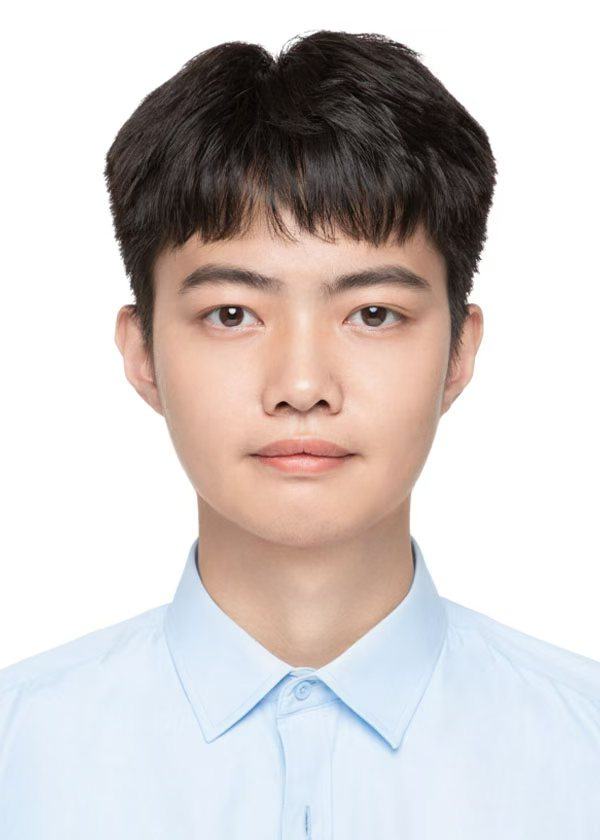}}]{Guangrui Bai}
Guangrui Bai received the M.S. degree from the School of Engineering Science, University of Science and Technology of China (USTC) in 2023. He is currently pursuing the Ph.D. degree at USTC. His research interests include low-light image enhancement and robotic vision.
\end{IEEEbiography}
\vspace{-40pt}

\begin{IEEEbiography}[{\includegraphics[width=1in,height=1.25in,clip,keepaspectratio]{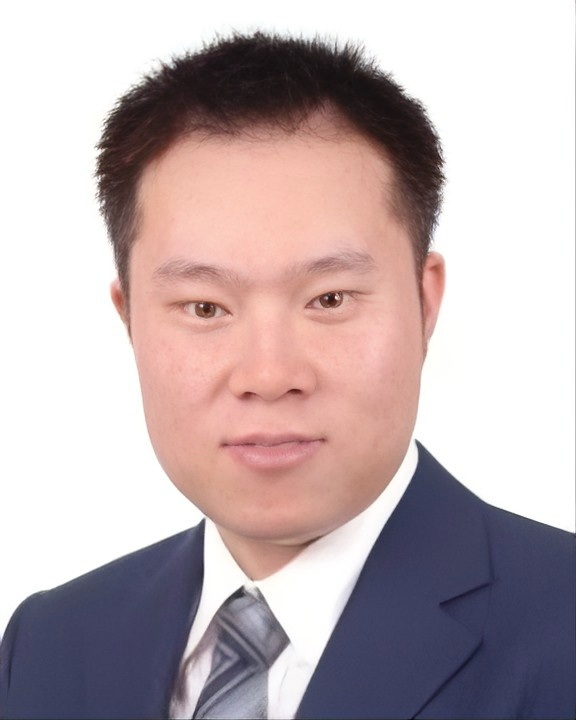}}]{Wenbo Chu}
Wenbo Chu received his B.S. degree majored in Automotive Engineering from Tsinghua University, China, in 2008, and his M.S. degree majored in Automotive Engineering from RWTH-Aachen, German and Ph.D. degree majored in Mechanical Engineering from Tsinghua University, China, in 2014. He is currently a research fellow at Western China Science City Innovation Center of Intelligent and Connected Vehicles (Chongqing) Co, Ltd., and National Innovation Center of Intelligent and Connected Vehicles.
\end{IEEEbiography}
\vspace{-40pt}

\begin{IEEEbiography}[{\includegraphics[width=1in,height=1.25in,clip,keepaspectratio]{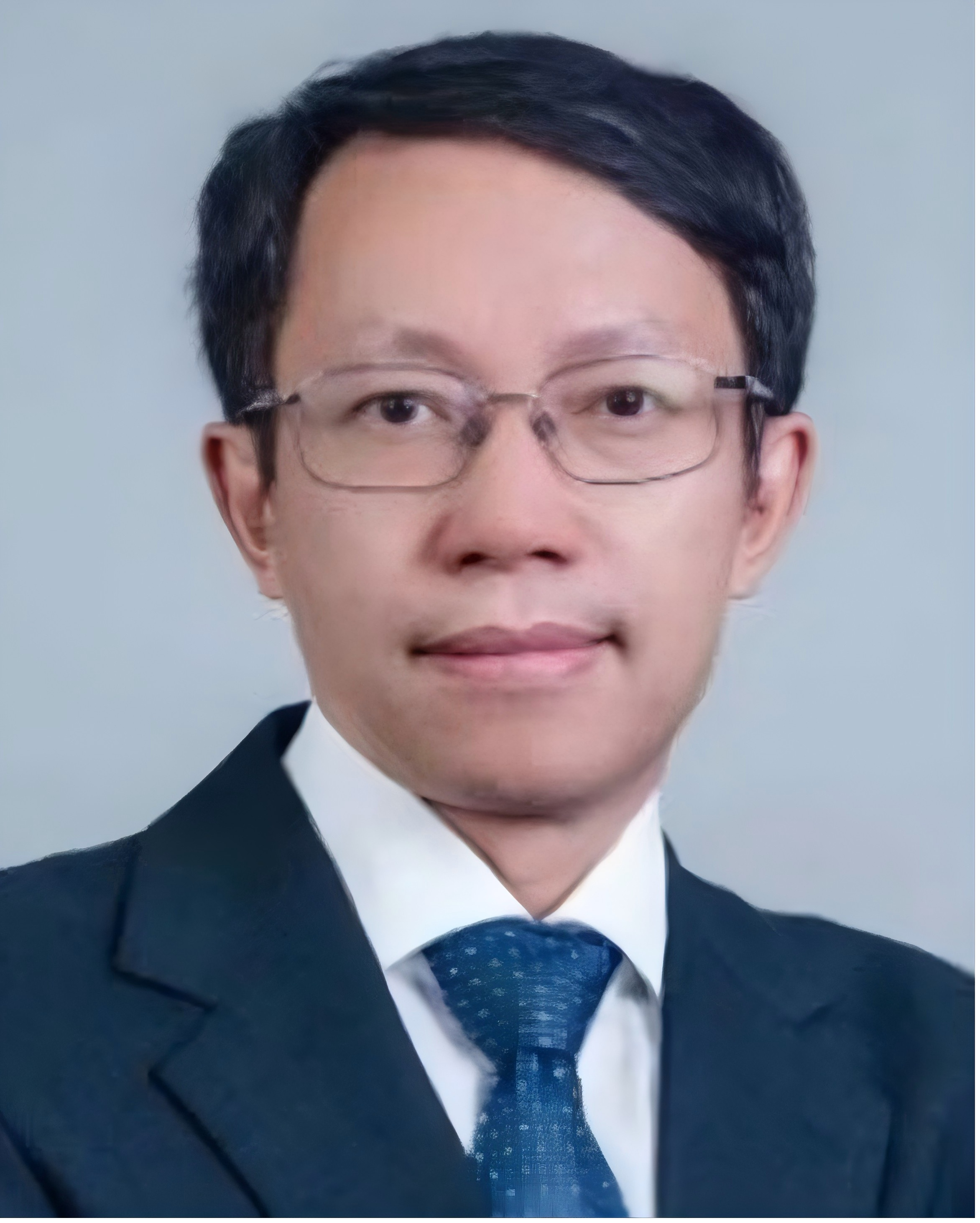}}]{Keqiang Li}
Keqiang Li received the B.E. degree from Tsinghua University, Beijing, China, in 1985, and the M.E. and Ph.D. degrees from Chongqing University, Chongqing, China, in 1988 and 1995, respectively. He is currently a Professor with the School of Vehicle and Mobility, Tsinghua University. He is the Chief Scientist of Intelligent and Connected Vehicle Innovation Center of China, and the Director of State Key Laboratory of Automotive Safety and Energy of China. His current research interests include intelligent connected vehicles, cloud-based control for vehicles, and vehicle dynamics systems.
\end{IEEEbiography}
\end{document}